\documentclass[sn-mathphys-num]{sn-jnl}%

\usepackage{amssymb}%
\usepackage{pifont}%
\usepackage{multirow}
\usepackage{xcolor}
\usepackage{algorithm}
\usepackage{algorithmic}
\usepackage{amsmath}
\usepackage{booktabs}
\usepackage{soul}
\usepackage{graphicx}
\usepackage{booktabs}
\usepackage{adjustbox}
\usepackage{array}
\usepackage{multirow}
\usepackage{multirow}
\usepackage{tabulary}
\usepackage{tabularx}
\usepackage[normalem]{ulem}
\newcommand{\xmark}{\ding{55}}%
\newcommand*\AlgCommentInLine[1]{{\color{blue}{$\triangleright$ \textit{#1}}}}

\begin{document}

\title[Article Title]{COBRA: A Continual Learning Approach to Vision-Brain Understanding}

\author*[1]{\fnm{Xuan-Bac} \sur{Nguyen}}\email{xnguyen@uark.edu}
\author[1]{\fnm{Manuel}  \sur{Serna-Aguilera}} \email{mserna@uark.edu}
\author[2]{\fnm{Arabinda Kumar Choudhary}} \email{choudhaa@upstate.edu}

\author[3]{\fnm{Pawan} \sur{Sinha}}\email{psinha@mit.edu}
\author[4]{\fnm{Xin} \sur{Li}}\email{xli48@albany.edu}
\author[1]{\fnm{Khoa} \sur{Luu}}\email{khoaluu@uark.edu}

\affil*[1]{\orgdiv{Dept. of Electrical Engineering and Computer Science}, \orgname{University of Arkansas}, \state{AR}, \country{US}}
\affil*[2]{\orgdiv{Dept. of Radiology}, \orgname{SUNY Upstate Medical University}, \state{NY}, \country{US}}
\affil*[3]{\orgdiv{Dept. of Brain \& Cognitive Sciences}, \orgname{Massachusetts Institute of Technology}, \state{MA}, \country{US}}
\affil*[4]{\orgdiv{Dept. of Computer Science}, \orgname{SUNY Albany}, \state{NY}, \country{US}}

\newcommand{\Bac}[1]{\textcolor{black}{#1}}
\newcommand{\BacRebuttal}[1]{\textcolor{black}{#1}}

\abstract{Vision-Brain Understanding (VBU) aims to extract visual information perceived by humans from brain activity recorded through functional Magnetic Resonance Imaging (fMRI). Despite notable advancements in recent years, existing studies in VBU continue to face the challenge of catastrophic forgetting, where models lose knowledge from prior subjects as they adapt to new ones. Addressing continual learning in this field is, therefore, essential. This paper introduces a novel framework called Continual Learning for Vision-Brain (COBRA) to address continual learning in VBU. Our approach includes three novel modules: a Subject Commonality (SC) module, a Prompt-based Subject Specific (PSS) module, and a transformer-based module for fMRI, denoted as MRIFormer module. The SC module captures shared vision-brain patterns across subjects, preserving this knowledge as the model encounters new subjects, thereby reducing the impact of catastrophic forgetting. On the other hand, the PSS module learns unique vision-brain patterns specific to each subject. Finally, the MRIFormer module comprises a transformer encoder and decoder that learn the fMRI features for VBU from both common and specific patterns. In a continual learning setup, COBRA is trained on new PSS and MRIFormer modules for new subjects, while the modules for previous subjects remain unaffected. As a result, COBRA effectively addresses catastrophic forgetting and achieves state-of-the-art performance in both continual learning and vision-brain reconstruction tasks, surpassing previous methods.
}


\maketitle

\section{Introduction}\label{sec:intro}

\begin{figure}[!ht]
    \centering
    \includegraphics[width=1.0\linewidth]{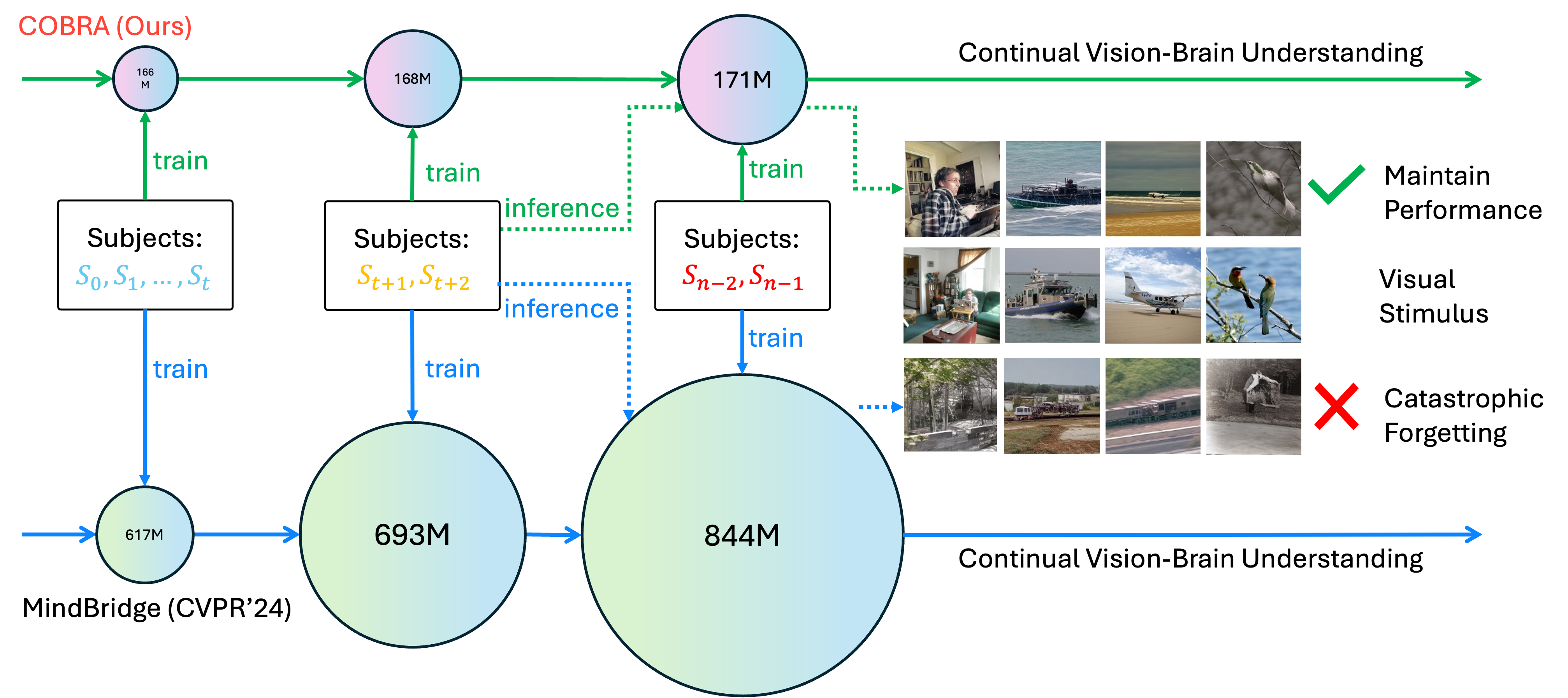}
    \caption{The paper introduces a novel framework, COBRA, for continual vision-brain understanding. In this figure, the sizes of the circles represent the network sizes. During the continual learning process, COBRA gradually increases the network size, whereas the previous method, MindBridge \cite{mindbridge}, significantly expands the model size. Furthermore, COBRA maintains performance, whereas MindBridge suffers from catastrophic forgetting. \textbf{Best viewed in color}}
    \label{fig:abstract_figure}
\end{figure}

Recently, vision-brain understanding (VBU) has made substantial progress due to advances in generative models, including GANs \cite{ozcelik2022reconstruction, shen2019deep} and Diffusion models \cite{takagi, mindeye1, mindbridge, psychometry, neuropictor, mindeye2}. These approaches allow the reconstruction of increasingly realistic and accurate images from functional Magnetic Resonance Imaging (fMRI) signals. The approaches designed per-model-per-subject, as seen in \cite{takagi, mindeye1}, are required to train a new model for a new subject. It can learn subject-specific but ignore subject-commonality features. To address this limitation, the studies \cite{mindbridge, psychometry, neuropictor, mindeye2} proposed unified models trained on multiple subjects simultaneously. While these approaches are trained on a close set of subjects, they may still face a degradation in performance on new subjects. Recent efforts \cite{mindbridge, mindeye2} have aimed to adapt pre-trained models to new subject data, yet they continue to suffer from performance drops on prior subjects due to catastrophic forgetting \cite{thrun1998lifelong, french1999catastrophic, robins1995catastrophic}, where models lose learned knowledge when adapting to new subject shifts. The ideal solution to this problem is to update the entire model with prior and new data. However, given the sensitive nature of brain data and the importance of privacy, retaining previous training data is often infeasible. Moreover, as the number of subjects grows, computational training costs become a concern. Thus, in practice, VBU models need to continually learn from new subjects without requiring retraining on prior data. This paradigm is known as Continual Vision-Brain Understanding (CVBU). Given the limitation in previous brain understanding approaches, we raise critical questions for CVBU: (1) \textit{What types of knowledge can the model retain}, and (2) \textit{what types of knowledge must it learn specifically from new subject data?}

Since all subjects share similar patterns of brain activations \cite{downing2001cortical, costantini2011haptic, epstein1998cortical, weiner2017cytoarchitecture, epstein1999parahippocampal, epstein2019scene, duchaine2015revised, haxby1996face, haxby1995hemispheric, grill2017functional}, if the model can understand these patterns, there is no need to learn from the data of new subjects. Consequently, the answer to the first question is to retain the knowledge of common patterns of brain activations learned from prior subjects and apply them to new ones. To achieve this, we propose a novel module designed to capture these patterns. Unlike prior methods \cite{downing2001cortical, costantini2011haptic, epstein1998cortical, weiner2017cytoarchitecture, epstein1999parahippocampal, nguyen2024quantum, nguyen2024quantumencode, truong2025insect, epstein2019scene, duchaine2015revised, haxby1996face, haxby1995hemispheric, grill2017functional}, our proposed approach can automatically identify a broader set of common patterns.

In addition to shared patterns of brain activations, each subject also has unique patterns. Therefore, in the continual brain understanding paradigm, the model must be able to adapt to these subject-specific patterns. This capability provides the answer to the second question. To achieve this, we draw inspiration from recent advancements in prompt-based learning or prompting \cite{liu2023pre}. This technique enables the model to receive learnable prompt tokens that flexibly incorporate additional task-specific information. In our approach, we treat subject-specific patterns as prompts representing unique information, which can then be learned and integrated from new subject data. This promises to mitigate the catastrophic forgetting problem in CVBU effectively.

\noindent
\textbf{Contributions of this Work:} In summary, this work presents a novel continual learning approach to VBU. The contributions of this work can be summarized as follows. 
First, we introduce a new Continual Learning for Vision-Brain Understanding (COBRA) framework to the visual-brain understanding problem. To the best of our knowledge, this is the first work to tackle the vision-brain problem using a continual learning approach (detailed in Fig. \ref{fig:abstract_figure}).
Second, a new Subject Commonality (SC) module will be presented to capture shared brain activation patterns across subjects. Unlike prior methods, COBRA can automatically identify a broader range of common patterns. 
Third, we present a new Prompt-based Subject-Specific (PSS) module to capture unique patterns specific to each subject. The module is trained to form the most relevant prompt, representing each subject’s distinct characteristics. 
Fourth, we introduce a novel MRIFormer module that contains a transformer encoder-decoder to construct fMRI features for VBU. Apart from prior methods, this module can explore sequence information, being flexible and independent of the length of target features, i.e., Contrastive Language-Image Pre-Training (CLIP) features. 
Fifth, we introduce a simple yet effective training strategy to address the CVBU problem. In particular, COBRA is only required to update the PSS and MRIFormer modules for new subjects. It leaves previous subjects’ PSS and MRIFormer unaffected, thus addressing catastrophic forgetting. Notably, the PSS module training is \textit{label-free}, eliminating the need for additional annotations. The extensive experiments demonstrate that our method achieves state-of-the-art (SOTA) performance in VBU and continual learning. The pros and cons of COBRA compared to prior methods are summarized in Table \ref{tab:summary}.

\begin{table}
    \centering
    \caption{Summary of the advantages and limitations of the previous methods. \textit{UM}: Unified model, \textit{SCL}: Subject Commonality Learning, \textit{SSL}: Subject Specific Learning, \textit{CL}: Continual Learning}
    \setlength{\tabcolsep}{8pt}
    \label{tab:summary}
    \begin{tabular}{l|cccc}
        \hline
         Method & UM & SCL & SSL & CL \\
         \hline
         Takagi \cite{takagi} & \xmark & \xmark & \checkmark & \xmark \\
         MindEye1 \cite{mindeye1} & \xmark & \xmark & \checkmark & \xmark  \\
         Psychometry \cite{mindeye2} & \checkmark & \checkmark & \checkmark & \xmark  \\
         MindBridge \cite{mindbridge} & \checkmark & \checkmark & - & \xmark \\
         MindEye2 \cite{mindeye2} & \checkmark & \checkmark & - & \xmark \\
         NeuroPictor \cite{neuropictor} & \checkmark & \checkmark & - & \xmark  \\
         UMBRAE \cite{xia-umbrae-2024} & \checkmark & \checkmark & - & \xmark  \\
         \hline
         COBRA &\checkmark & \checkmark & \checkmark & \checkmark \\
         \hline
    \end{tabular}
\end{table}

\section{Related Work}\label{sec:formatting}

\subsection{Vision-Brain Understanding}
In VBU, the fMRI signals corresponding to subjects' viewing of an image are treated as input, which is then decoded to reconstruct the image originally viewed \cite{neuropictor, mindeye1, mindeye2, psychometry, mindbridge, takagi, xia-umbrae-2024}. 
Many works have leveraged diffusion \cite{thermodynamics_2015, ho-denoising-2020, openai-diffusion-2021, rombach-highresolutionimagesynthesislatent-2022, xu-versatilediffusiontextimages-2024} to learn the distribution of the fMRI signal. 
MindEye \cite{mindeye1} aligns the original image and fMRI latent spaces via contrastive learning and then uses diffusion to reconstruct the original image stimulus. 
MindEye2 \cite{mindeye2} trains in a limited data setting to address expensive fMRI data acquisition. %
Takagi et al. \cite{takagi} use latent diffusion \cite{rombach-highresolutionimagesynthesislatent-2022} for image reconstruction, using a part of the fMRI as the diffusion conditioning. 
MindBridge \cite{mindbridge} trains with a cyclic fMRI reconstruction mechanism, aligning brain data across subjects, which enables more robust brain-to-image and text decoding using dual embeddings. 
Psychometry \cite{psychometry}, by contrast, trains a singular model with common and unique features for all subjects. 
NeuroPictor \cite{neuropictor} uses a latent diffusion-like approach \cite{rombach-highresolutionimagesynthesislatent-2022} to encode the fMRI signals into a latent space to separate high- and low-level features, giving the ability to focus on finer details.
UMBRAE \cite{xia-umbrae-2024} decodes fMRI signals in a vision-language framework, aligning text descriptions and image features.

\subsection{Continual Learning}
We review two continual learning approaches \cite{wang-learningpromptcontinuallearning-2022}, each with advantages and limitations.
Rehearsal-based methods address catastrophic forgetting by storing sampled past tasks’ data in buffers \cite{hayes-memoryefficientexperiencereplay-2019, chaudhry-tinyepisodicmemoriescontinual-2019, chaudhry-usinghindsightanchorpast-2021, chaudhry-efficientlifelonglearningagem-2019, buzzega-darkexperiencegeneralcontinual-2020, rebuffi-icarlincrementalclassifierrepresentation-2017, reza-privacy-dl-2015, pham-dualnetcontinuallearningfast-2021, cha-co2lcontrastivecontinuallearning-2021, wu-largescaleincrementallearning-2019}, and thus integrate old and new data to prevent knowledge loss.
Several works leverage knowledge distillation to compress knowledge across older and newer tasks \cite{chaudhry-usinghindsightanchorpast-2021, buzzega-darkexperiencegeneralcontinual-2020, rebuffi-icarlincrementalclassifierrepresentation-2017, wu-largescaleincrementallearning-2019} while others leverage self-supervised methods \cite{pham-dualnetcontinuallearningfast-2021, cha-co2lcontrastivecontinuallearning-2021,nguyen2019audio,nguyen2019sketch,nguyen2020self,nguyen2021clusformer,nguyen2022multi,nguyen2022two,nguyen2023algonauts,nguyen2023brainformer,nguyen2023fairness,nguyen2024bractive,nguyen2024diffusion,nguyen2024hierarchical,nguyen2024insect,nguyen2024qclusformer}.
Key limitations include buffer size, where making it too small degrades learning \cite{cha-co2lcontrastivecontinuallearning-2021} and data privacy, restricting access to past data \cite{reza-privacy-dl-2015}.
Architecture-based methods address catastrophic forgetting by modifying the architecture itself.
Typically, this is done by adding a new set of parameters for new tasks \cite{zhao-deepbayesianunsupervisedlifelong-2021, yoon-lifelonglearningdynamicallyexpandable-2019, rusu-progressiveneuralnetworks-2022, rao-continualunsupervisedrepresentationlearning-2019, loo-generalizedvariationalcontinuallearning-2020, li-learngrowcontinualstructure-2019} or sub-networks specialized for a certain task are maintained \cite{wortsman-supermaskssuperposition-2020, serra-overcomingcatastrophicforgettinghard-2018, mallya-packnetaddingmultipletasks-2018, ke-continuallearningmixedsequence-2021}.
Key limitations include the added complexity of added parameters and the model needing the task type beforehand to select the task parameters, which is not guaranteed to be given at test time.

\subsection{Limitations of Data Representation in Prior Methods}
\label{sec:limitations}
\begin{figure}[!t]
    \centering
    \includegraphics[width=1\linewidth]{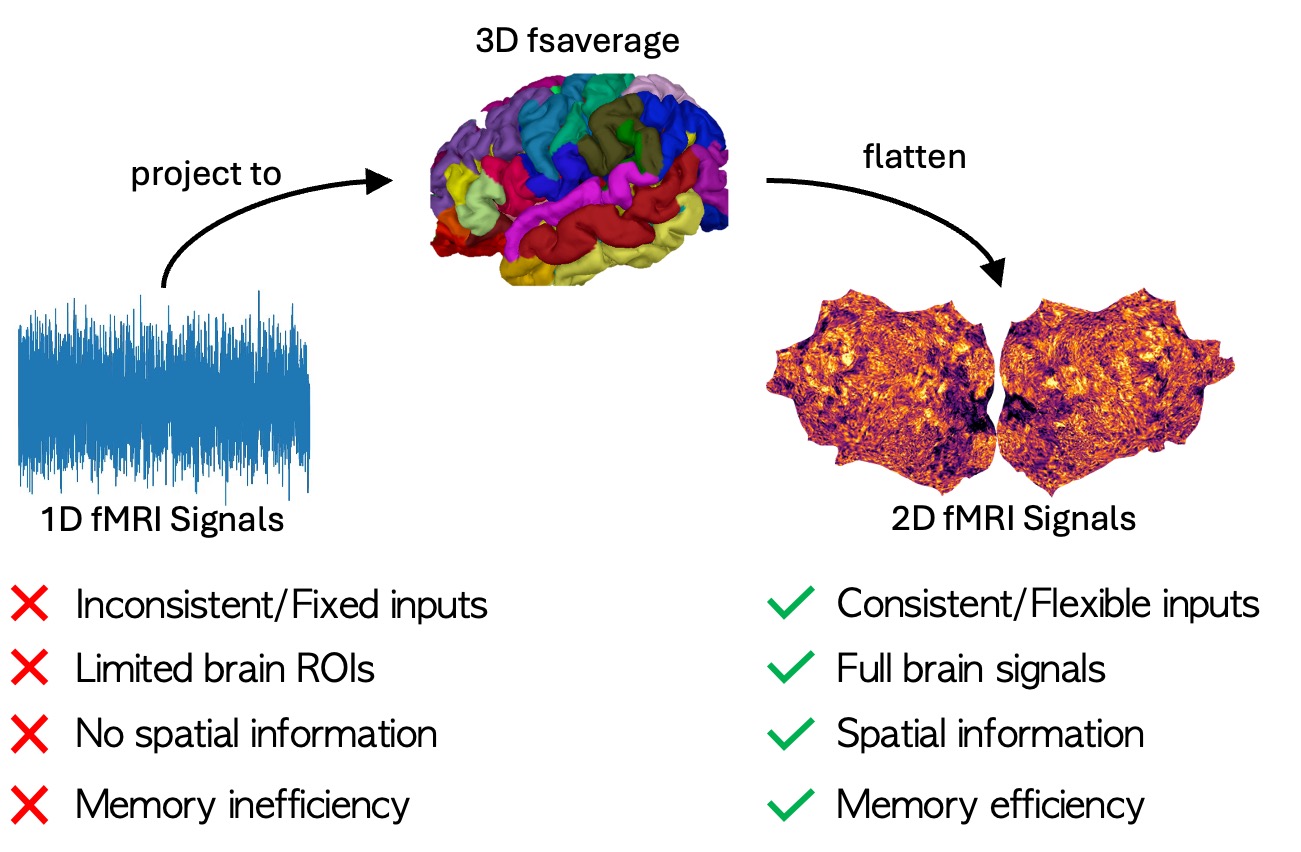}
    \caption{Limitations of previous work and our advancements in vision-brain representation.}
    \label{fig:data_review}
\end{figure}

From prior studies, we have identified three main limitations, particularly concerning continual brain understanding and the unified model approaches.
\newline\noindent
\textbf{L1.} Since the brain sizes and structures of each subject differ, the sizes of fMRI signals vary across subjects. These signal sizes range from approximately 13,000 to 16,000 voxels per subject. Prior methods \cite{defossez2023decoding, yu2021mixup, liang2022mind, tan2019efficientnet, caron2021emerging, haxby2001distributed, takagi, mindeye1, psychometry, mindbridge, mindeye2, neuropictor} \textit{fixed the input signal size} and adapted to new subjects with different input sizes. This poses a significant challenge to understanding the brain continually, requiring methods to work with inputs of varying dimensions.
\newline\noindent
\textbf{L2.} Treating brain signals as one-dimensional (1D) vectors leads to a loss of spatial information. While voxels are closely connected to their neighbors in 3D space, flattening the signals into 1D disrupts these interactions. In particular, we observed that two samples in the signals may not be adjacent to each other spatially in 3D space, making it difficult to interpret the connections between them.
\newline\noindent
\textbf{L3.} Prior methods \textit{rely on signals from specific Regions of Interest (ROI)}, such as floc-face, floc-body, and floc-place, rich in information about human visual perception. However, this focus may overlook information about other neighborhoods. For instance, the floc-face area activates when a subject views a face in a visual stimulus, while the floc-body is responsible for perceiving the human body. This approach may fail when applied to non-human subjects. To the best of our knowledge, no specific ROIs are dedicated to such common objects. Therefore, using signals from only the face, body, place, and word ROIs might miss critical information related to other elements in the visual stimulus.

\section{The Proposed COBRA Approach}

In this section, we first formulate the continual learning for vision-brain understanding in Section \ref{sec:continual_brain_decoding}. Next, we present a data preprocessing approach as in Section \ref{subsec:revisit_data_representation} to address essential conditions so that our proposed modules, i.e., Subject Commonality (Section \ref{subsec:sc}) and Prompt-based Subject Specific Modules (Section \ref{sec:pss}), can address introduced limitations \textbf{L1, L2, L3} as mentioned in Section \ref{sec:limitations}.

\subsection{Continual Vision-Brain Understanding}
\label{sec:continual_brain_decoding}

In this section, we first introduce continual brain understanding protocols. While typical CL
is usually defined as training machine learning models on non-stationary data from sequential tasks, CVBU shares a similar idea. However, instead of increasing the tasks for the model as in CL, CVBU increases the number of subjects added to the most recent models. 

We define a sequence of brain understanding tasks: $D = \{D_{1} \dots  D_{S} \}$, where the $s^{th}$ subject $D_{s} = \{(v_i, x_i)\}_{i=1}^{n_s}$ is a set of pairs where each pair contains a visual stimulus $v_i \in \mathcal{V}$ and its corresponding record fMRI signals $x_i \in \mathcal{X}$. $n_s$ is the number of samples for the task, and $i$ is the sample index of the task dataset. The goal \Bac{of subject incremental learning in VBU} is to train a model $\mathcal{M}_{\theta}: \mathcal{X} \rightarrow \mathcal{V}$ to predict $v = \mathcal{M}_{\theta}(x) \in \mathcal{V}$ for \Bac{minimizing the total risk of all subjects}. 
\begin{align}
    \mathcal{L}^*(\theta) = \mathcal{L}_s(\theta) + \mathcal{L}_{1:s-1}(\theta) = \mathbb{E}_{(x, v) \sim \mathcal{D}_s}\left[\mathcal{L}(\mathcal{M}_{\theta}(x), v)\right] + \sum_{i=1}^{s-1} \mathbb{E}_{(x, v) \sim \mathcal{D}_i}\left[\mathcal{L}(\mathcal{M}_{\theta}(x), v)\right]
\end{align}
\Bac{where $\mathcal{L}_s$ calculates the expected loss error $\mathcal{L}$ of $\mathcal{M}_{\theta}$ while training with subject $s^{th}$. $\mathcal{L}_{1:s-1}$ is the total error evaluated on the past subjects trained on the previous steps. The challenge in the CVBU is that data points from previous subjects may not be available during training on new subjects due to several reasons, such as privacy concerns. This paper aims to tackle this problem by learning from new subjects while maintaining the performance of past subjects. 
\BacRebuttal{In traditional task-incremental continual learning, tasks differ by objective (e.g., classification of different categories), however, in CVBU, each subject shares the same objective, e.g., reconstructing or decoding the same image space. The CVBU setup, however, introduces a unique distribution shift due to individual brain structure, neural responses, and voxel patterns.} 
\BacRebuttal{This makes each subject a non-independent and identically distributed (non-i.i.d. distribution) over the same task space rather than a new task. Moreover, CVBU has unique constraints: (1) no access to past subject data due to privacy, (2) high-dimensional, structured, and spatially variable input (fMRI), and (3) shared decoder space (CLIP features), unlike task-specific heads used in task-incremental CL}. The details of the proposed method will be presented in the next sections.} 

\subsection{Data Preprocessing}
\label{subsec:revisit_data_representation}

To address the limitations of \textbf{L1}, we adopt the 3D FreeSurfer \cite{fischl1999high} (fsaverage) space to represent the fMRI signals in the 2D forms. Since the 3D FreeSurfer is a general template of the human brain, it can effectively represent the fMRI signals of all subjects.
In addition, the 3D FreeSurfer template offers 3D spatial information, which can help to address \textbf{L2}.
While \textbf{L3} will be majorly addressed via our Subject Commonality Module (presented in the next section), we have found that the data preprocessing step plays an important role in providing the essential conditions to address \textbf{L3}.
In particular, instead of using fMRI signals from specific ROIs, we use the entire brain signal, hypothesizing that the model will learn to identify the relevant regions for each ROI, automatically.
In conclusion, representing brain signals in 2D addresses the previous limitations and offers flexibility for various tasks, such as continuous brain understanding and unified model learning.
It provides an advanced approach to understanding the brain by leveraging CNNs or transformers.
Fig. \ref{fig:data_review} (right-hand side) illustrates the representation of the 2D form of the fMRI signals. \Bac{It is important to note that, the 2D form of fMRI signals does not lead to information loss since data transformations are just purely projection operations.}
\BacRebuttal{In particular, the 2D representation used in COBRA is derived from flattening the 3D cortical surface (fsaverage) into 2D, preserving geodesic distance and spatial relationships, similar to how FreeSurfer is used in neuroscience.
Importantly, this is a bijective projection. Each voxel has a unique mapping from 3D to 2D and vice versa, with no downsampling or interpolation. Thus, the original spatial resolution is maintained. To further support this, in the Appendix Sec \ref{subsec:appendix_spatial_issue}, we explain more about this issue.}
For convenience, in the rest of the paper, we refer to fMRI signals in the 2D flattened form instead of in 1D.

\begin{figure*}
    \centering
    \includegraphics[width=1\linewidth]{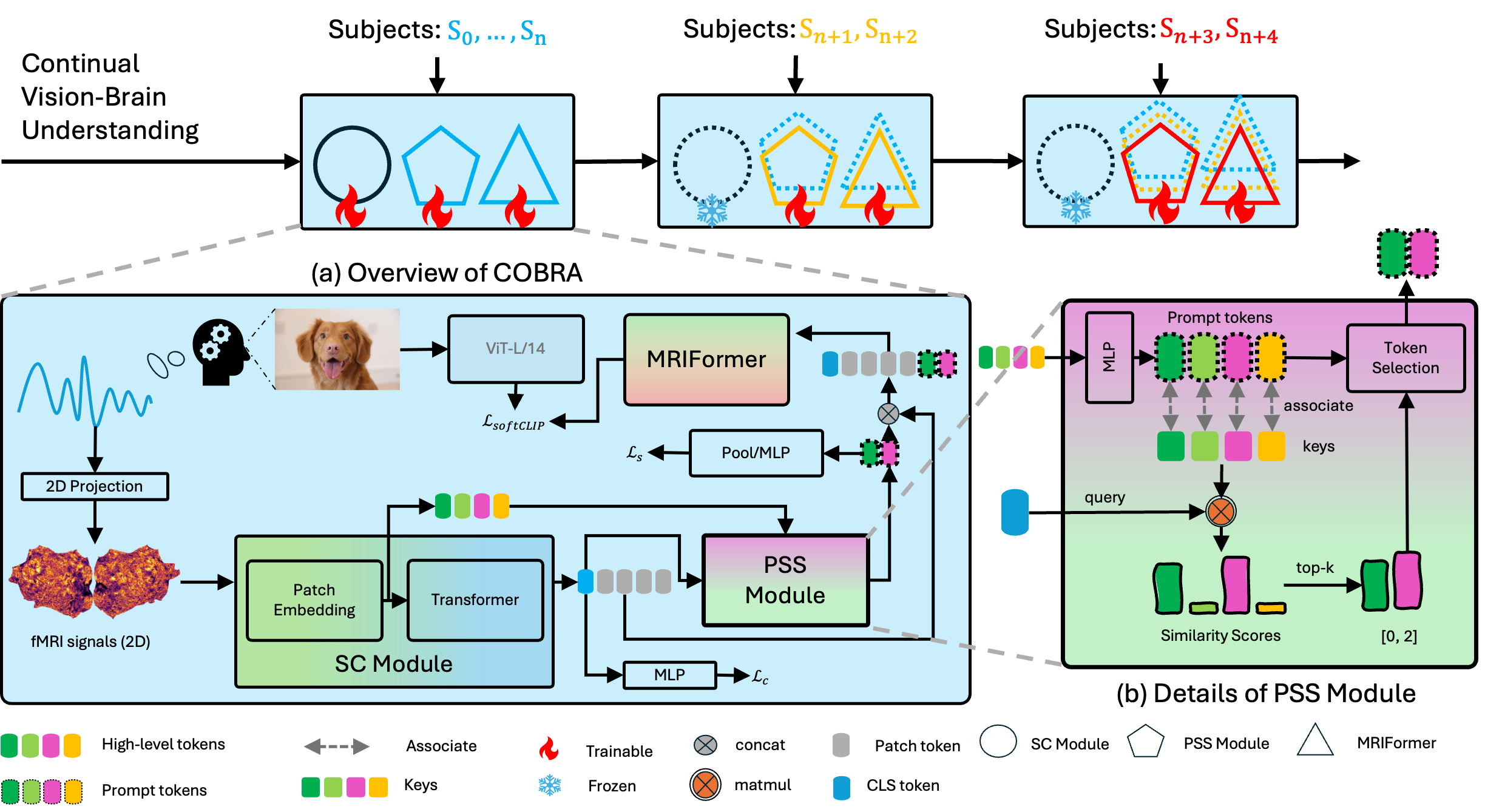}
    \caption{
    \textbf{Overview of the COBRA Framework}. COBRA comprises three modules: Subject Commonality (SC), Prompt-based Subject Specific (PSS), and MRIFormer. The SC module is designed to capture common vision-brain patterns across subjects, while the PSS module focuses on learning subject-specific patterns. The MRIFormer module is a transformer-based architecture, consisting of an encoder that integrates common and subject-specific features, and a decoder that transforms the fMRI feature into a unified CLIP space.
    }
    \label{fig:proposed_framework}
\end{figure*}

\subsection{Subject Commonality Module}
\label{subsec:sc}

This section introduces the Subject Commonality (SC) module, designed to capture shared features across subjects.
This module is inspired by neuroscience research to identify the functional localizers (floc) of brain patterns related to specific visual stimuli \cite{downing2001cortical, costantini2011haptic, epstein1998cortical, weiner2017cytoarchitecture, epstein1999parahippocampal, epstein2019scene, duchaine2015revised, haxby1996face, haxby1995hemispheric, grill2017functional}.
This module can also address \textbf{L3} and discover ROIs of different categories rather than face, body, etc. 

Given the fMRI signals $x \in \mathbb{R}^{H \times W \times C}$ represented in 2D format, we define SC module as a transformer-based network, denoted as $\mathcal{M}_{c}$, to extract the commonality feature $f_c = \left[f_c^{\texttt{CLS}}, f_p \right] = \mathcal{M}_{c}(x) \in \mathbb{R}^{(L_c + 1) \times D}$ where $f_p \in \mathbb{R}^{L_c \times D}$ contains high-level patch tokens which are output of Patch Embedding module, $L_c$ is number of patches. $f_c^{\texttt{CLS}} \in \mathbb{R}^D$ denoted as $\texttt{CLS}$ token. The goal of this module is to find out what objects fMRI signals represent. The output of the SC module is formed as follows,

\begin{equation}
\begin{split}
    \label{eq:common_features}
    f_c &= \left[f_c^{\texttt{CLS}}, f_p \right] = \mathcal{M}_{c}(x) \in \mathbb{R}^{L_c \times D} \\
    \hat{y_c} &= \texttt{MLP}_{c}(f_c^{\texttt{CLS}}) \in \mathbb{R}^{N_c} 
\end{split}
\end{equation}
where $\texttt{MLP}_{c}$ is a linear layer, $\hat{y_c}$ is the predicted objects in the fMRI signals, $N_c$ is number of object. In the Natural Scenes Dataset \cite{allen2022massive} (NSD), $N_c$ is fixed to 80 since all stimuli are from the COCO \cite{lin2014microsoft} database. In addition, each stimulus contains multiple objects; for that reason, the $\mathcal{M}_{c}(x)$ is optimized using Binary Cross Entropy as in the Eqn \eqref{eq:cs_loss}. 

\begin{equation}
    \label{eq:cs_loss}
    \mathcal{L}_c = -\sum_{i=0}^{N_c-1} (y_{c,i} \times  \log(\hat{y}_{c,i}) + (1 -  y_{c,i}) \times \log(1 - \hat{y}_{c,i}))
\end{equation}
\noindent
Since $\mathcal{M}_{c}$ is designed to receive the signal from any subjects,  $\mathcal{M}_{c}$ will learn \textit{common} representation across subjects. Secondly, $\mathcal{M}_{c}$ aims to learn the context inside the signals, i.e., what objects the signal contains.

\subsection{Prompt-based Subject Specific (PSS) Module}
\label{sec:pss}

\noindent
\textbf{Challenges}. 
Let $x^{(a,v)}$ and $x^{(b,v)}$ denote the fMRI signals of $a^{th}$ subject and $b^{th}$ subject that are perceiving visual stimulus $v \in \mathcal{V}$. It is obvious that $x^{(a,v)} \neq x^{(b,v)}$ since they are collected from different persons. Since both subjects are seeing the same stimulus $v$, we can learn how different their brain patterns are by training a classification model $\mathcal{M}_{s}$ to predict the signal $x^{v} \in \{x^{(a,v)}, x^{(b,v)}\}$ belonging to either subject $a^{th}$ or $b^{th}$. In NSD, all subjects share 1,000 visual stimuli during recording. However, these samples belong to the testing set only, so we could not utilize them for training.
                
\noindent
\textbf{Motivations}. 
Even if all subjects see different visual stimuli, they still experiment on the same set of objects $\mathcal{C} = \{\text{dog, cat, car, truck, \dots.}\}$. Specifically, for the NSD database, $\mathcal{C}$ is a set of 80 COCO classes. Instead of learning specific features across subjects on the same visual stimulus strictly, we learn subject-specific features via a similar context, i.e., $f_c^{\texttt{CLS}}$ that subjects are viewing.

\noindent
To achieve this goal, we draw inspiration from recent advances in prompt-based learning, also known as prompting \cite{liu2023pre, wang2022dualprompt, wang2022learning}, a novel transfer learning technique in natural language processing (NLP). 
Prompting techniques design a model to generate textual inputs with templated or learnable prompt tokens containing additional task-specific information. In this paper, we treat prompts as subject-specific features. We need to form a prompt that describes this unique aspect.

\noindent
\textbf{Methods}. 
First, we need to build a pool of prompt tokens that contain the semantic information of fMRI signals. Fortunately, we found that the high-level tokens $f_p$ can fulfill these requirements. Let $\mathcal{P} = \texttt{MLP}_p(f_p) \in \mathbb{R}^{L_c \times D}$ be prompt tokens where $\texttt{MLP}_p$ is a linear layer. Let $\mathcal{K} \in \mathbb{R}^{L_c \times D}$ be a set of keys where each key will be associated with a token. To construct a prompt, we design the key-query paradigm to select the most $k$-suitable tokens w.r.t $f_c^{\texttt{CLS}}$ as in the Eqn \eqref{eq:specific_features}. 
\begin{equation}\label{eq:specific_features}
\begin{split}
    \text{sim} &= f_c^{\texttt{CLS}} \otimes \mathcal{K}^{T} \in \mathbb{R}^{L_c}   \\
    \text{index} &= \texttt{TopK}(\text{sim}, \text{k})   \\
    f_s &= \texttt{IndexSelect}(\mathcal{P}, \text{index}) \in \mathbb{R}^{k \times D}
\end{split}
\end{equation}
where $\otimes$ denotes a matrix multiplication operation, $f_s$ is the subject-specific feature. We feed $f_s$ into a pooling layer followed by a linear layer and then classify the feature corresponding to the right subjects as in Eqn \eqref{eqn:pss_loss}.
\begin{equation}
\label{eqn:pss_loss}
\begin{split}
    \hat{y}_s &= \texttt{MLP}_{s}(\texttt{AvgPool}(f_s)) \in \mathbb{R}^{N_s}   \\
    \mathcal{L}_{s} &= -\sum_{i=0}^{N_s-1} y_{s,i} \times  \log(\hat{y}_{s,i})
\end{split}
\end{equation}
where $N_s$ is the number of subjects. Interestingly, the training SSL module is \textit{label-free} since all we need to know is the subject identity. 

\subsection{MRIFormer}
With the subject common feature $f_c$ from Eqn \eqref{eq:common_features} and the subject-specific feature $f_s$ from Eqn \eqref{eq:specific_features}, we concatenate both to get the feature of fMRI signals as follows,
\begin{equation}
    f = \texttt{concat}(f_c, f_s) \in \mathbb{R}^{(L_c + k) \times D}
\end{equation}
The feature $f$ contains multiple tokens from fMRI images and subject-specific prompts. We further feed $f$ into a transformer encoder to accumulate information between commonalities and specific features as in Eqn \eqref{eq:encoder}. 
\begin{equation}
    \label{eq:encoder}
    f_h = \texttt{TransEnc}(f) \in \mathbb{R}^{(L_c + k) \times D}
\end{equation}
The feature $f_h$ will be aligned with the corresponding Contrastive Language-Image Pre-Training (CLIP) feature of the visual stimulus: $f_{CLIP} = \mathcal{M}_{CLIP}(v) \in \mathbb{R}^{L_{CLIP} \times D}$. We observe an inconsistency in the length of the feature, i.e., $(L_c + k) \neq L_{CLIP}$. Prior studies \cite{takagi, mindeye1, mindbridge, mindeye2} utilized a linear layer to map the fMRI feature to the same length as the CLIP one. However, it will lose the sequence properties of a feature. To address this problem, we take inspiration from Neural Machine Translation to design a transformer decoder to \textit{translate} fMRI features to CLIP features. This approach can preserve sequential information and address feature length inconsistency, as shown in Eqn. \eqref{eq:decoder}. 

\begin{align}
\label{eq:decoder}
    f_{mri} = \texttt{TransDecoder}(f_h, f_{q}) \in \mathbb{R}^{L_{CLIP} \times D}
\end{align}
where $f_{q} \in \mathbb{R}^{L_{CLIP} \times D}$ is a query vector. Finally, the $f_{mri}$ feature is aligned with $f_{CLIP}$ using Contrastive Loss \cite{clip-paper-2021} as in Eqn \eqref{eqn:contrastive}.

\begin{equation} \label{eqn:contrastive}
\begin{split}
    \mathcal{L}_{con} &= -\frac{1}{N}\sum_i^N \left[\log\frac{\exp(\mathbf{p}_i \otimes \mathbf{q}_i/\sigma)}{\sum_j^N\exp(\mathbf{p}_i \otimes \mathbf{q}_j/\sigma)} - \log\frac{\exp(\mathbf{q}_i \otimes \mathbf{p}_i/\sigma)}{\sum_j^N\exp(\mathbf{q}_i \otimes \mathbf{p}_j/\sigma)}\right]
\end{split}
\end{equation}
where $\textbf{p} = f_{fmri}$, $\textbf{q} = f_{CLIP}$ and $\sigma$ is the temperature.

\subsection{Continual Training Strategy}
As outlined in Section \ref{sec:continual_brain_decoding}, CVBU involves updating models to accommodate new subject data without forgetting knowledge from previously trained subjects. 
Assuming the SC module has acquired substantial knowledge from previous subjects, it focuses on learning subject-specific features of each new subject. Therefore, for each new subject added to the pipeline, we do not retrain or update the entire model. Instead, we only need to create new PSS and MRIFormer modules and train them to capture features specific to the new subject. Fig. \ref{fig:proposed_framework} illustrates the pipeline of CVBU.

As in the CVBU, the size of $\texttt{MLP}_s$ in Eqn \eqref{eqn:pss_loss} will be extended as the number of subjects increases. This layer holds \textit{center vectors} of the subjects. Since we do not have access to previous subjects, the center vectors of previous subjects are not updated in the current steps of adapting new subjects. Therefore, the new center vector may be collapsed into the existing center vectors, making the PSS module fail to learn the specific feature. To address this problem, we add a regulation loss as follows, 
\begin{equation}
    \mathcal{L}_{reg} = \sum_{c_i, c_j} \{\text{max}\left(0, 2\nabla - ||c_i - c_j||\right)\}^2
\end{equation}
where $c_i, c_j$ are the center vectors of $i^{th}$ and $j^{th}$ subjects and $\nabla$ is the margin between centers.

\noindent
\textbf{Training Loss Functions.}
In summary, COBRA is trained using the objective loss function as follows,
\begin{equation}
\label{eqn:total_loss}
    \mathcal{L} = \lambda_c \mathcal{L}_c + \lambda_s \mathcal{L}_s + \lambda_{sc} \mathcal{L}_{{con}} + \lambda_{reg} \mathcal{L}_{reg}
\end{equation}
where $\lambda_c, \lambda_s, \lambda_{sc}, \lambda_{reg}$ are the weights of the SC module, PSS module, Soft CLIP, and regulation loss, respectively. The pseudo-code for COBRA is described in Algorithm \ref{algo:COBRA}.

\begin{figure}[!t]
\vspace{-1.5mm}
\begin{algorithm}[H]
\centering
\footnotesize
\caption{Pseudo code for COBRA.}
\label{algo:COBRA}
\begin{algorithmic}[1]
\STATE {\bfseries Input:} The visual stimulus $v$, fMRI signals $x$
\STATE {\bfseries Output:} The objective loss.

\AlgCommentInLine{Commonality Module}
\STATE{$f_c \gets \left[f_c^{\texttt{CLS}}, f_p \right] \gets \mathcal{M}_c$}(x)
\STATE{$\hat{y}_c \gets \texttt{MLP}_c(f_c^{\texttt{CLS}})$}
\STATE{$\mathcal{L}_c \gets -\sum_{i=0}^{N_c-1} (y_{c,i} \times  \log(\hat{y}_{c,i}) + (1 -  y_{c,i}) \times \log(1 - \hat{y}_{c,i}))$}

\AlgCommentInLine{Prompt-based Subject Specific Module}
\STATE{$\mathcal{P} \gets \texttt{MLP}_p(f_p)$}
\STATE{$\text{sim} \gets f_c^{\texttt{CLS}} \otimes \mathcal{K}^{T}$}
\STATE{$\text{index} \gets \texttt{TopK}(\text{sim}, \text{k})$}
\STATE{$f_s \gets \texttt{IndexSelect}(\mathcal{P}, \text{index})$}
\STATE{$\hat{y}_s \gets \texttt{MLP}_{s}(\texttt{AvgPool}(f_s))$}
\STATE{$\mathcal{L}_{s} \gets -\sum_{i=0}^{N_s-1} y_{s,i} \times  \log(\hat{y}_{s,i})$}
\STATE{$ \mathcal{L}_{reg} \gets \sum_{c_i, c_j} \{\text{max}\left(0, 2\nabla - ||c_i - c_j||\right)\}^2$}

\AlgCommentInLine{MRIFormer}
\STATE{$f \gets \texttt{Concat}(f_c, f_s)$}
\STATE{$f_h \gets \texttt{TransEnc}(f)$}
\STATE{$f_{mri} \gets \texttt{TransDecoder}(f_h, f_{q})$}
\STATE{$f_{CLIP} \gets \mathcal{M}_{CLIP}(v)$}
\STATE{$\mathcal{L}_{{con}} \gets \texttt{ContrastiveLoss}(f_{mri}, f_{CLIP})$}

\AlgCommentInLine{Training loss}
\STATE{$\mathcal{L} \gets \lambda_c \mathcal{L}_c + \lambda_s \mathcal{L}_s + \lambda_{sc} \mathcal{L}_{\text{softCLIP}}+ \lambda_{reg} \mathcal{L}_{reg}$}
\RETURN{$\mathcal{L}$}
\end{algorithmic}
\end{algorithm}
\vspace{-10mm}
\end{figure}

\subsection{Implementation Details}
The 2D fMRI signals $x$ are resized to $224 \times 224$. We select $\texttt{vit-base-16}$ as the SC module to extract the patches' features of the fMRI signals. Each feature vector has dimension $D=768$. In the PSS module, we design associated keys for prompt tokens as a vector of $197 \times 768$. We select $k=30$ as the top $k$ number. That means for each query of commonality feature $f_c^{\texttt{CLS}}$, we select top-30 tokens from the high-level tokens $\mathcal{P}$ to describe specific features of the subject.
The transformer encoder in Eqn \eqref{eq:encoder} and the decoder in Eqn. \eqref{eq:decoder} is designed to have four in-depth, dimension dim is 768, and the number of heads is set to 12. The framework is implemented using PyTorch and trained on 16 $\times$ A100 GPUs (40GB each). The learning rate is set to $2.5e^{-5}$ initially and then reduced to zero gradually under ConsineLinear \cite{martino2020semeval} policy. The batch size is set to $32$/GPU. The weight factors are set to one equally in the Eqn \ref{eqn:total_loss}.
The model is optimized by AdamW \cite{loshchilov2017decoupled} for $300$ epochs. The training is completed within approximately two hours per subject.

\section{Experimental Results}

\subsection{Dataset and Protocol}

\textbf{Dataset.} Following common practices \cite{defossez2023decoding, yu2021mixup, liang2022mind, tan2019efficientnet, caron2021emerging, haxby2001distributed}, we utilize the Natural Scenes Dataset \cite{allen2022massive} for the brain understanding task. This dataset includes high-resolution 7-Tesla fMRI scans collected from eight healthy subjects who viewed thousands of natural images from MS-COCO \cite{lin2014microsoft}.

\noindent
\textbf{Protocol.}
We propose two different CL scenarios.
First, in the {(3,4)-(6,8)-(1,2)-(5,7)} setup, training begins with subjects (3,4), followed by subjects (6,8), (1,2), and (5,7) sequentially, for a total of four training steps.
Second, in the (3, 4, 6, 8)-(1, 2, 5, 7) setup, the model is trained on subjects (3, 4, 6, 8) first, followed by subjects (1, 2, 5, 7), for a total of two training steps.
To evaluate the CL performance, we use the final model from each training scenario and test it on the test sets of subjects (3, 4, 6, 8) and (1, 2, 5, 7). \BacRebuttal{We begin with the (3,4,6,8) groups (incomplete subjects), then add (1,2,5,7) (complete subjects) to simulate a realistic scenario where earlier data is limited, and richer data becomes available over time. In addition, we report performance separately on the (3,4,6,8) and (1,2,5,7) groups to enable fair comparisons between incomplete and complete data cohorts. This also allows for direct performance comparison with prior works, which often report results on the (1,2,5,7) set.} We also evaluate the performance of these models on all subjects.

\noindent
\textbf{Metrics.} \BacRebuttal{For vision-brain reconstruction task, }similar to prior studies \cite{defossez2023decoding, yu2021mixup, liang2022mind, tan2019efficientnet, caron2021emerging, haxby2001distributed}, 
we adopt the Structural Similarity Index (SSIM) and Contrastive Language-Image Pre-Training (CLIP) score to evaluate the low-level and high-level reconstruction results. \BacRebuttal{For brain caption task, similar to prior studies \cite{xia-umbrae-2024}, we employ BLEU-1, BLEU-2, BLEU-3, BLEU-4, METEOR, ROUGE, CIDEr, SPICE, CLIP-S, and RefCLIP-S to evaluate the caption results}. 

\begin{figure*}
    \centering
    \includegraphics[width=0.85\linewidth]{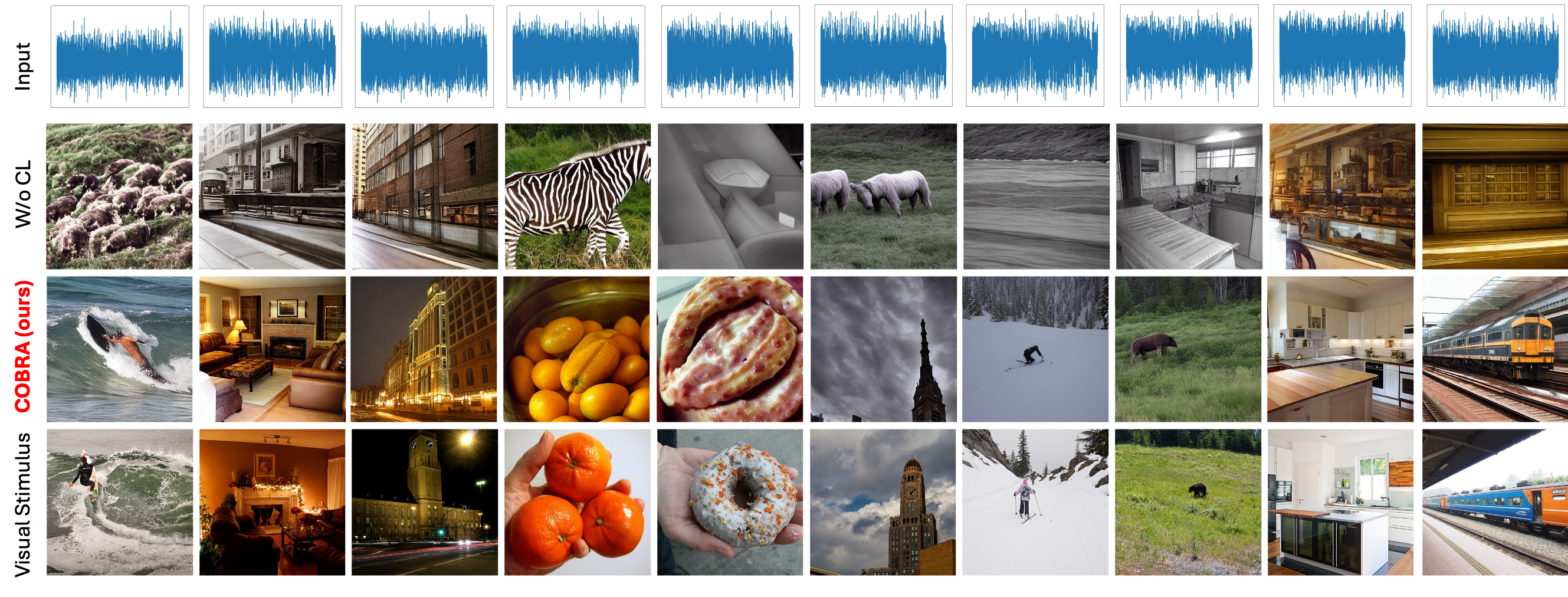}
    \caption{Qualitative results of the proposed COBRA in continual vision-brain understanding task. The first and top row is the input fMRI signals. The second row is the results without continual learning. The third row is the results of COBRA. The last row is the stimulus.}
    \label{fig:cl_qualitative_results}
\end{figure*}

\subsection{Rehearsal-free Continual Learning} \label{subsec:RefreeCon}
\subsubsection{Rehearsal-free Vision-Brain Reconstruction Task}
\label{subsubsec:vision-brain-recon}
In this benchmark, we do not use any samples from the previous steps to train at the current step. The performance is presented in the Table \ref{tab:cl-main}. Without continual learning (W/o CL), performance drops significantly to 0.207 for the Structural Similarity Index (SSIM) and 59.8\% for CLIP. Our COBRA approach achieves the scores of 0.320 for SSIM and 93.2\% for CLIP, on average. It outperforms previous methods, such as PLOP \citep{plop} and LwF \citep{lwf}, by 0.04 and 0.073 for SSIM, and by 17.2\% and 25.8\% for CLIP, respectively. In the {(3,4,6,8)-(1,2,5,7)} setup, we observed similar results. COBRA achieves the scores of 0.329 for SSIM and 92.9 for CLIP, on average. It outperforms PLOP by 0.037 for SSIM and 14.8\% for CLIP and LwF by 0.08 for SSIM and 28.6\% for CLIP.
Interestingly, we found that our COBRA can cooperate with the PLOP \cite{plop} method. By doing so, COBRA achieves higher performance, i.e., 0.01 for SSIM and 1\% for CLIP in the 4-step scenario, and 0.004 for SSIM and 0.8\% for CLIP in the 2-step scenario.

\begin{table}[t]
\centering
\caption{
   Rehearsal-free continual learning results in the NSD Testing Set. The bold numbers indicate the best results, while the underlines indicate the second-best results. The asterisk ($*$) indicates the results of our COBRA in cooperating with PLOP \cite{plop}. The dagger ($^\dagger$) indicates the results of our COBRA trained on all subjects.
}
\label{tab:cl-main}
\setlength{\tabcolsep}{2pt}
\begin{tabular}{|l|cc|cc|cc|}
\hline
\multicolumn{7}{|c|}{(3,4)-(6,8)-(1,2)-(5,7) (4 steps)} \\
\hline
\multirow{2}{*}{Method}
& \multicolumn{2}{c|}{(3,4,6,8)}
& \multicolumn{2}{c|}{(1,2,5,7)}
& \multicolumn{2}{c|}{all} \\
\cmidrule{2-7}
& SSIM & CLIP & SSIM & CLIP & SSIM & CLIP \\
\hline
W/o CL & 0.197{\tiny$\pm$0.005} & 58.9{\tiny$\pm$0.6} & 0.216{\tiny$\pm$0.006} & 60.6{\tiny$\pm$0.7} & 0.207{\tiny$\pm$0.005} & 59.8{\tiny$\pm$0.6} \\
LwF \cite{lwf} & 0.233{\tiny$\pm$0.006} & 69.1{\tiny$\pm$0.5} & 0.277{\tiny$\pm$0.007} & 65.6{\tiny$\pm$0.5} & 0.255{\tiny$\pm$0.007} & 67.4{\tiny$\pm$0.4} \\
PLOP \cite{plop} & 0.268{\tiny$\pm$0.005} & 75.8{\tiny$\pm$0.4} & 0.308{\tiny$\pm$0.005} & 76.2{\tiny$\pm$0.4} & 0.288{\tiny$\pm$0.005} & 76.0{\tiny$\pm$0.4} \\
COBRA & \underline{0.311{\tiny$\pm$0.004}} & \underline{93.2{\tiny$\pm$0.3}} & \underline{0.344{\tiny$\pm$0.004}} & \underline{93.1{\tiny$\pm$0.3}} & \underline{0.328{\tiny$\pm$0.004}} & \underline{93.2{\tiny$\pm$0.3}} \\
COBRA$^*$ & \textbf{0.325{\tiny$\pm$0.003}} & \textbf{94.4{\tiny$\pm$0.2}} & \textbf{0.351{\tiny$\pm$0.003}} & \textbf{94.8{\tiny$\pm$0.2}} & \textbf{0.338{\tiny$\pm$0.003}} & \textbf{94.2{\tiny$\pm$0.2}} \\
\textcolor{gray}{COBRA$^\dagger$} & \textcolor{gray}{0.336{\tiny$\pm$0.002}} & \textcolor{gray}{96.9{\tiny$\pm$0.2}} & \textcolor{gray}{0.365{\tiny$\pm$0.002}} & \textcolor{gray}{97.2{\tiny$\pm$0.2}} & \textcolor{gray}{0.351{\tiny$\pm$0.002}} & \textcolor{gray}{97.1{\tiny$\pm$0.2}} \\
\hline
\multicolumn{7}{|c|}{(3,4,6,8)-(1,2,5,7) (2 steps)} \\
\hline
\multirow{2}{*}{Method}
& \multicolumn{2}{c|}{(3,4,6,8)}
& \multicolumn{2}{c|}{(1,2,5,7)}
& \multicolumn{2}{c|}{all} \\
\cmidrule{2-7}
& SSIM & CLIP & SSIM & CLIP & SSIM & CLIP \\
\hline
W/o CL & 0.197{\tiny$\pm$0.005} & 59.7{\tiny$\pm$0.6} & 0.238{\tiny$\pm$0.007} & 61.1{\tiny$\pm$0.6} & 0.218{\tiny$\pm$0.006} & 60.4{\tiny$\pm$0.6} \\
LwF \cite{lwf} & 0.235{\tiny$\pm$0.006} & 63.8{\tiny$\pm$0.5} & 0.263{\tiny$\pm$0.006} & 64.7{\tiny$\pm$0.4} & 0.249{\tiny$\pm$0.006} & 64.3{\tiny$\pm$0.4} \\
PLOP \cite{plop} & 0.272{\tiny$\pm$0.005} & 76.8{\tiny$\pm$0.4} & 0.311{\tiny$\pm$0.006} & 79.1{\tiny$\pm$0.5} & 0.292{\tiny$\pm$0.005} & 78.0{\tiny$\pm$0.5} \\
COBRA & \underline{0.316{\tiny$\pm$0.004}} & \underline{92.3{\tiny$\pm$0.3}} & \underline{0.342{\tiny$\pm$0.004}} & \underline{93.4{\tiny$\pm$0.3}} & \underline{0.329{\tiny$\pm$0.004}} & \underline{92.9{\tiny$\pm$0.3}} \\
COBRA$^*$ & \textbf{0.322{\tiny$\pm$0.003}} & \textbf{93.5{\tiny$\pm$0.2}} & \textbf{0.345{\tiny$\pm$0.003}} & \textbf{93.8{\tiny$\pm$0.2}} & \textbf{0.333{\tiny$\pm$0.003}} & \textbf{93.7{\tiny$\pm$0.2}} \\
\textcolor{gray}{COBRA$^\dagger$} & \textcolor{gray}{0.336{\tiny$\pm$0.002}} & \textcolor{gray}{96.9{\tiny$\pm$0.2}} & \textcolor{gray}{0.365{\tiny$\pm$0.002}} & \textcolor{gray}{97.2{\tiny$\pm$0.2}} & \textcolor{gray}{0.351{\tiny$\pm$0.002}} & \textcolor{gray}{97.1{\tiny$\pm$0.2}} \\
\hline
\end{tabular}
\end{table}

\subsubsection{Rehearsal-free Brain Captioning Task}
\label{subsubsec:brain-captioning}
\BacRebuttal{Building on the models trained for the Vision-Brain Reconstruction task described in Section~\ref{subsubsec:vision-brain-recon} and following the approach in~\cite{xia-umbrae-2024}, we employ a linear projector or adapter to map the fMRI feature representations into the language embedding space. These transformed features are then passed to a large language model (LLM), specifically Shikra~\cite{chen2023shikra}, to generate captions corresponding to brain signals. We evaluate the performance of rehearsal-free continual learning for this brain captioning task across multiple metrics: BLEU-1 and BLEU-2 in Table~\ref{tab:caption_bleu1_bleu2}, BLEU-3 and BLEU-4 in Table~\ref{tab:caption_bleu3_bleu4}, METEOR and ROUGE in Table~\ref{tab:caption_meteor_rouge}, CIDEr and SPICE in Table~\ref{tab:caption_cider_spice}, and CLIP-S and RefCLIP-S in Table~\ref{tab:caption_clips}. Overall, the performance trends are consistent with those observed in the vision-brain reconstruction task. Across most metrics and both continual learning settings (4-step and 2-step), our proposed method, COBRA, consistently outperforms competing approaches, including LwF~\cite{lwf}, PLOP~\cite{plop}, and the baseline without continual learning by a substantial margin. These results further highlight the robustness and effectiveness of COBRA across diverse vision-brain understanding tasks.}

\begin{table}[!ht]
\centering
\caption{
Rehearsal-free continual learning results on NSD caption using BLEU-1 and BLEU-2 metrics (\%). The bold numbers indicate the best results, while the underlines indicate the second-best results. Results are reported as mean $\pm$ std.
}
\label{tab:caption_bleu1_bleu2}
\begin{tabular}{|l|cc|cc|cc|}
\hline
\multicolumn{7}{|c|}{(3,4)-(6,8)-(1,2)-(5,7) (4 steps)} \\
\hline
Method & \multicolumn{2}{c|}{(3,4,6,8)} & \multicolumn{2}{c|}{(1,2,5,7)} & \multicolumn{2}{c|}{all} \\
\cmidrule{2-7}
& BLEU-1 & BLEU-2 & BLEU-1 & BLEU-2 & BLEU-1 & BLEU-2 \\
\hline
W/o CL     & 47.2{\tiny$\pm$0.4} & 31.2{\tiny$\pm$0.3} & 49.1{\tiny$\pm$0.4} & 32.9{\tiny$\pm$0.3} & 48.1{\tiny$\pm$0.4} & 32.0{\tiny$\pm$0.3} \\
LwF \cite{lwf}        & 51.0{\tiny$\pm$0.3} & 34.0{\tiny$\pm$0.3} & 52.3{\tiny$\pm$0.3} & 35.1{\tiny$\pm$0.3} & 51.6{\tiny$\pm$0.3} & 34.6{\tiny$\pm$0.3} \\
PLOP \cite{plop}       & 53.3{\tiny$\pm$0.3} & 35.7{\tiny$\pm$0.3} & 54.6{\tiny$\pm$0.3} & 36.9{\tiny$\pm$0.3} & 53.9{\tiny$\pm$0.3} & 36.3{\tiny$\pm$0.3} \\
COBRA      & \underline{58.6{\tiny$\pm$0.2}} & \underline{39.8{\tiny$\pm$0.2}} & \underline{59.4{\tiny$\pm$0.2}} & \underline{40.5{\tiny$\pm$0.2}} & \underline{59.0{\tiny$\pm$0.2}} & \underline{40.2{\tiny$\pm$0.2}} \\
COBRA$^*$  & \textbf{59.2{\tiny$\pm$0.2}} & \textbf{40.4{\tiny$\pm$0.2}} & \textbf{60.2{\tiny$\pm$0.2}} & \textbf{41.2{\tiny$\pm$0.2}} & \textbf{59.7{\tiny$\pm$0.2}} & \textbf{40.9{\tiny$\pm$0.2}} \\
\textcolor{gray}{COBRA$^\dagger$}
           & \textcolor{gray}{62.6{\tiny$\pm$0.2}} & \textcolor{gray}{41.7{\tiny$\pm$0.2}} & \textcolor{gray}{62.0{\tiny$\pm$0.2}} & \textcolor{gray}{43.1{\tiny$\pm$0.2}} & \textcolor{gray}{62.3{\tiny$\pm$0.2}} & \textcolor{gray}{42.4{\tiny$\pm$0.2}} \\
\hline
\multicolumn{7}{|c|}{(3,4,6,8)-(1,2,5,7) (2 steps)} \\
\hline
Method & \multicolumn{2}{c|}{(3,4,6,8)} & \multicolumn{2}{c|}{(1,2,5,7)} & \multicolumn{2}{c|}{all} \\
\cmidrule{2-7}
& BLEU-1 & BLEU-2 & BLEU-1 & BLEU-2 & BLEU-1 & BLEU-2 \\
\hline
W/o CL     & 47.5{\tiny$\pm$0.4} & 31.6{\tiny$\pm$0.3} & 48.9{\tiny$\pm$0.4} & 32.7{\tiny$\pm$0.3} & 48.2{\tiny$\pm$0.4} & 32.2{\tiny$\pm$0.3} \\
LwF \cite{lwf}        & 51.2{\tiny$\pm$0.3} & 34.2{\tiny$\pm$0.3} & 52.6{\tiny$\pm$0.3} & 35.5{\tiny$\pm$0.3} & 51.9{\tiny$\pm$0.3} & 34.9{\tiny$\pm$0.3} \\
PLOP \cite{plop}       & 53.6{\tiny$\pm$0.3} & 36.0{\tiny$\pm$0.3} & 54.8{\tiny$\pm$0.3} & 37.1{\tiny$\pm$0.3} & 54.2{\tiny$\pm$0.3} & 36.6{\tiny$\pm$0.3} \\
COBRA      & \underline{59.8{\tiny$\pm$0.2}} & \underline{41.0{\tiny$\pm$0.2}} & \underline{60.8{\tiny$\pm$0.2}} & \underline{41.8{\tiny$\pm$0.2}} & \underline{60.3{\tiny$\pm$0.2}} & \underline{41.4{\tiny$\pm$0.2}} \\
COBRA$^*$  & \textbf{60.6{\tiny$\pm$0.2}} & \textbf{41.8{\tiny$\pm$0.2}} & \textbf{61.5{\tiny$\pm$0.2}} & \textbf{42.6{\tiny$\pm$0.2}} & \textbf{61.1{\tiny$\pm$0.2}} & \textbf{42.2{\tiny$\pm$0.2}} \\
\textcolor{gray}{COBRA$^\dagger$}
           & \textcolor{gray}{62.6{\tiny$\pm$0.2}} & \textcolor{gray}{41.7{\tiny$\pm$0.2}} & \textcolor{gray}{62.0{\tiny$\pm$0.2}} & \textcolor{gray}{43.1{\tiny$\pm$0.2}} & \textcolor{gray}{62.3{\tiny$\pm$0.2}} & \textcolor{gray}{42.4{\tiny$\pm$0.2}} \\
\hline
\end{tabular}
\end{table}

\begin{table}[!ht]
\centering
\caption{Rehearsal-free continual learning results on NSD caption using BLEU-3 and BLEU-4 metrics (\%). The bold numbers indicate the best results, while the underlines indicate the second-best results. Results are reported as mean $\pm$ std.
}
\label{tab:caption_bleu3_bleu4}
\begin{tabular}{|l|cc|cc|cc|}
\hline
\multicolumn{7}{|c|}{(3,4)-(6,8)-(1,2)-(5,7) (4 steps)} \\
\hline
Method & \multicolumn{2}{c|}{(3,4,6,8)} & \multicolumn{2}{c|}{(1,2,5,7)} & \multicolumn{2}{c|}{all} \\
\cmidrule{2-7}
& BLEU-3 & BLEU-4 & BLEU-3 & BLEU-4 & BLEU-3 & BLEU-4 \\
\hline
W/o CL     & 19.3{\tiny$\pm$0.4} & 11.6{\tiny$\pm$0.3} & 20.1{\tiny$\pm$0.5} & 12.1{\tiny$\pm$0.2} & 19.7{\tiny$\pm$0.3} & 11.9{\tiny$\pm$0.4} \\
LwF \cite{lwf}        & 21.7{\tiny$\pm$0.3} & 13.3{\tiny$\pm$0.2} & 22.6{\tiny$\pm$0.4} & 13.9{\tiny$\pm$0.3} & 22.1{\tiny$\pm$0.2} & 13.6{\tiny$\pm$0.4} \\
PLOP \cite{plop}       & 23.0{\tiny$\pm$0.2} & 14.3{\tiny$\pm$0.5} & 24.1{\tiny$\pm$0.4} & 15.1{\tiny$\pm$0.3} & 23.5{\tiny$\pm$0.3} & 14.7{\tiny$\pm$0.3} \\
COBRA      & \underline{26.1}{\tiny$\pm$0.4} & \underline{16.4}{\tiny$\pm$0.3} & \underline{27.2}{\tiny$\pm$0.5} & \underline{17.1}{\tiny$\pm$0.3} & \underline{26.7}{\tiny$\pm$0.4} & \underline{16.8}{\tiny$\pm$0.2} \\
COBRA$^*$  & \textbf{26.7{\tiny$\pm$0.5}} & \textbf{16.9{\tiny$\pm$0.4}} & \textbf{27.8{\tiny$\pm$0.3}} & \textbf{17.5{\tiny$\pm$0.3}} & \textbf{27.3{\tiny$\pm$0.4}} & \textbf{17.2{\tiny$\pm$0.3}} \\
\textcolor{gray}{COBRA$^\dagger$} & 
\textcolor{gray}{29.6{\tiny$\pm$0.3}} & \textcolor{gray}{19.1{\tiny$\pm$0.2}} &
\textcolor{gray}{31.0{\tiny$\pm$0.3}} & \textcolor{gray}{20.5{\tiny$\pm$0.2}} &
\textcolor{gray}{30.3{\tiny$\pm$0.3}} & \textcolor{gray}{19.8{\tiny$\pm$0.2}} \\
\hline
\multicolumn{7}{|c|}{(3,4,6,8)-(1,2,5,7) (2 steps)} \\
\hline
Method & \multicolumn{2}{c|}{(3,4,6,8)} & \multicolumn{2}{c|}{(1,2,5,7)} & \multicolumn{2}{c|}{all} \\
\cmidrule{2-7}
& BLEU-3 & BLEU-4 & BLEU-3 & BLEU-4 & BLEU-3 & BLEU-4 \\
\hline
W/o CL     & 19.5{\tiny$\pm$0.3} & 11.8{\tiny$\pm$0.4} & 20.4{\tiny$\pm$0.2} & 12.4{\tiny$\pm$0.5} & 20.0{\tiny$\pm$0.4} & 12.1{\tiny$\pm$0.2} \\
LwF \cite{lwf}        & 22.0{\tiny$\pm$0.3} & 13.5{\tiny$\pm$0.2} & 23.0{\tiny$\pm$0.4} & 14.1{\tiny$\pm$0.3} & 22.6{\tiny$\pm$0.5} & 13.9{\tiny$\pm$0.3} \\
PLOP \cite{plop}       & 23.3{\tiny$\pm$0.4} & 14.6{\tiny$\pm$0.2} & 24.5{\tiny$\pm$0.3} & 15.4{\tiny$\pm$0.4} & 23.9{\tiny$\pm$0.2} & 15.0{\tiny$\pm$0.5} \\
COBRA      & \underline{27.4}{\tiny$\pm$0.3} & \underline{17.3}{\tiny$\pm$0.3} & \underline{28.3}{\tiny$\pm$0.2} & \underline{18.1}{\tiny$\pm$0.2} & \underline{27.9}{\tiny$\pm$0.4} & \underline{17.7}{\tiny$\pm$0.4} \\
COBRA$^*$  & \textbf{28.0{\tiny$\pm$0.5}} & \textbf{17.9{\tiny$\pm$0.4}} & \textbf{28.9{\tiny$\pm$0.3}} & \textbf{18.6{\tiny$\pm$0.3}} & \textbf{28.4{\tiny$\pm$0.4}} & \textbf{18.3{\tiny$\pm$0.3}} \\
\textcolor{gray}{COBRA$^\dagger$} & 
\textcolor{gray}{29.6{\tiny$\pm$0.3}} & \textcolor{gray}{19.1{\tiny$\pm$0.2}} &
\textcolor{gray}{31.0{\tiny$\pm$0.3}} & \textcolor{gray}{20.5{\tiny$\pm$0.2}} &
\textcolor{gray}{30.3{\tiny$\pm$0.3}} & \textcolor{gray}{19.8{\tiny$\pm$0.2}} \\
\hline
\end{tabular}
\end{table}

\begin{table}[!ht]
\centering
\caption{
Rehearsal-free continual learning results on NSD caption using METEOR and ROUGE metrics (\%). The bold numbers indicate the best results, while the underlines indicate the second-best results. Results are reported as mean $\pm$ std.
}
\label{tab:caption_meteor_rouge}
\begin{tabular}{|l|cc|cc|cc|}
\hline
\multicolumn{7}{|c|}{(3,4)-(6,8)-(1,2)-(5,7) (4 steps)} \\
\hline
Method & \multicolumn{2}{c|}{(3,4,6,8)} & \multicolumn{2}{c|}{(1,2,5,7)} & \multicolumn{2}{c|}{all} \\
\cmidrule{2-7}
& METEOR & ROUGE & METEOR & ROUGE & METEOR & ROUGE \\
\hline
W/o CL     & 14.2{\tiny$\pm$0.4} & 36.5{\tiny$\pm$0.2} & 15.1{\tiny$\pm$0.5} & 37.3{\tiny$\pm$0.3} & 14.6{\tiny$\pm$0.3} & 36.9{\tiny$\pm$0.4} \\
LwF \cite{lwf}        & 15.8{\tiny$\pm$0.2} & 38.1{\tiny$\pm$0.3} & 16.6{\tiny$\pm$0.4} & 39.0{\tiny$\pm$0.5} & 16.2{\tiny$\pm$0.3} & 38.6{\tiny$\pm$0.2} \\
PLOP \cite{plop}       & 17.1{\tiny$\pm$0.5} & 39.4{\tiny$\pm$0.4} & 17.9{\tiny$\pm$0.2} & 40.3{\tiny$\pm$0.2} & 17.5{\tiny$\pm$0.4} & 39.9{\tiny$\pm$0.3} \\
COBRA      & \underline{18.8}{\tiny$\pm$0.4} & \underline{42.0}{\tiny$\pm$0.3} & \underline{19.7}{\tiny$\pm$0.3} & \underline{42.8}{\tiny$\pm$0.4} & \underline{19.2}{\tiny$\pm$0.4} & \underline{42.4}{\tiny$\pm$0.3} \\
COBRA$^*$  & \textbf{19.3{\tiny$\pm$0.2}} & \textbf{42.5{\tiny$\pm$0.5}} & \textbf{20.3{\tiny$\pm$0.3}} & \textbf{43.3{\tiny$\pm$0.3}} & \textbf{19.8{\tiny$\pm$0.4}} & \textbf{42.9{\tiny$\pm$0.4}} \\
\textcolor{gray}{COBRA$^\dagger$} & 
\textcolor{gray}{21.7{\tiny$\pm$0.3}} & \textcolor{gray}{44.1{\tiny$\pm$0.2}} &
\textcolor{gray}{22.9{\tiny$\pm$0.3}} & \textcolor{gray}{45.4{\tiny$\pm$0.2}} &
\textcolor{gray}{22.3{\tiny$\pm$0.3}} & \textcolor{gray}{44.8{\tiny$\pm$0.2}} \\
\hline
\multicolumn{7}{|c|}{(3,4,6,8)-(1,2,5,7) (2 steps)} \\
\hline
Method & \multicolumn{2}{c|}{(3,4,6,8)} & \multicolumn{2}{c|}{(1,2,5,7)} & \multicolumn{2}{c|}{all} \\
\cmidrule{2-7}
& METEOR & ROUGE & METEOR & ROUGE & METEOR & ROUGE \\
\hline
W/o CL     & 14.5{\tiny$\pm$0.3} & 36.8{\tiny$\pm$0.3} & 15.4{\tiny$\pm$0.5} & 37.6{\tiny$\pm$0.4} & 15.0{\tiny$\pm$0.2} & 37.2{\tiny$\pm$0.2} \\
LwF \cite{lwf}        & 16.3{\tiny$\pm$0.4} & 38.4{\tiny$\pm$0.2} & 17.2{\tiny$\pm$0.5} & 39.3{\tiny$\pm$0.3} & 16.8{\tiny$\pm$0.3} & 38.9{\tiny$\pm$0.5} \\
PLOP \cite{plop}       & 17.6{\tiny$\pm$0.2} & 39.7{\tiny$\pm$0.4} & 18.5{\tiny$\pm$0.4} & 40.6{\tiny$\pm$0.3} & 18.1{\tiny$\pm$0.2} & 40.2{\tiny$\pm$0.5} \\
COBRA      & \underline{20.0}{\tiny$\pm$0.3} & \underline{43.1}{\tiny$\pm$0.5} & \underline{20.9}{\tiny$\pm$0.2} & \underline{43.9}{\tiny$\pm$0.2} & \underline{20.5}{\tiny$\pm$0.4} & \underline{43.5}{\tiny$\pm$0.3} \\
COBRA$^*$  & \textbf{20.5{\tiny$\pm$0.4}} & \textbf{43.6{\tiny$\pm$0.3}} & \textbf{21.4{\tiny$\pm$0.4}} & \textbf{44.4{\tiny$\pm$0.5}} & \textbf{21.0{\tiny$\pm$0.3}} & \textbf{44.0{\tiny$\pm$0.4}} \\
\textcolor{gray}{COBRA$^\dagger$} & 
\textcolor{gray}{21.7{\tiny$\pm$0.3}} & \textcolor{gray}{44.1{\tiny$\pm$0.2}} &
\textcolor{gray}{22.9{\tiny$\pm$0.3}} & \textcolor{gray}{45.4{\tiny$\pm$0.2}} &
\textcolor{gray}{22.3{\tiny$\pm$0.3}} & \textcolor{gray}{44.8{\tiny$\pm$0.2}} \\
\hline
\end{tabular}
\end{table}

\begin{table}[!ht]
\centering
\caption{
Rehearsal-free continual learning results on NSD caption using CIDEr and SPICE metrics (\%). The bold numbers indicate the best results, while the underlines indicate the second-best results. Results are reported as mean $\pm$ std.
}
\label{tab:caption_cider_spice}
\begin{tabular}{|l|cc|cc|cc|}
\hline
\multicolumn{7}{|c|}{(3,4)-(6,8)-(1,2)-(5,7) (4 steps)} \\
\hline
Method & \multicolumn{2}{c|}{(3,4,6,8)} & \multicolumn{2}{c|}{(1,2,5,7)} & \multicolumn{2}{c|}{all} \\
\cmidrule{2-7}
& CIDEr & SPICE & CIDEr & SPICE & CIDEr & SPICE \\
\hline
W/o CL     & 35.2{\tiny$\pm$0.4} & 6.8{\tiny$\pm$0.3} & 37.5{\tiny$\pm$0.5} & 7.3{\tiny$\pm$0.4} & 36.4{\tiny$\pm$0.2} & 7.1{\tiny$\pm$0.4} \\
LwF \cite{lwf}        & 40.8{\tiny$\pm$0.3} & 8.1{\tiny$\pm$0.5} & 43.0{\tiny$\pm$0.2} & 8.6{\tiny$\pm$0.3} & 41.9{\tiny$\pm$0.4} & 8.4{\tiny$\pm$0.2} \\
PLOP \cite{plop}       & 47.6{\tiny$\pm$0.3} & 8.8{\tiny$\pm$0.4} & 50.1{\tiny$\pm$0.4} & 9.5{\tiny$\pm$0.2} & 48.9{\tiny$\pm$0.3} & 9.2{\tiny$\pm$0.5} \\
COBRA      & \underline{56.4}{\tiny$\pm$0.5} & \underline{11.2}{\tiny$\pm$0.4} & \underline{58.7}{\tiny$\pm$0.3} & \underline{11.9}{\tiny$\pm$0.4} & \underline{57.6}{\tiny$\pm$0.2} & \underline{11.6}{\tiny$\pm$0.3} \\
COBRA$^*$  & \textbf{57.3{\tiny$\pm$0.3}} & \textbf{11.6{\tiny$\pm$0.2}} & \textbf{59.8{\tiny$\pm$0.4}} & \textbf{12.3{\tiny$\pm$0.5}} & \textbf{58.6{\tiny$\pm$0.3}} & \textbf{12.0{\tiny$\pm$0.4}} \\
\textcolor{gray}{COBRA$^\dagger$} &
\textcolor{gray}{62.0{\tiny$\pm$0.4}} & \textcolor{gray}{13.2{\tiny$\pm$0.3}} &
\textcolor{gray}{64.4{\tiny$\pm$0.4}} & \textcolor{gray}{13.9{\tiny$\pm$0.3}} &
\textcolor{gray}{63.2{\tiny$\pm$0.4}} & \textcolor{gray}{13.5{\tiny$\pm$0.3}} \\
\hline
\multicolumn{7}{|c|}{(3,4,6,8)-(1,2,5,7) (2 steps)} \\
\hline
Method & \multicolumn{2}{c|}{(3,4,6,8)} & \multicolumn{2}{c|}{(1,2,5,7)} & \multicolumn{2}{c|}{all} \\
\cmidrule{2-7}
& CIDEr & SPICE & CIDEr & SPICE & CIDEr & SPICE \\
\hline
W/o CL     & 36.0{\tiny$\pm$0.5} & 7.0{\tiny$\pm$0.2} & 38.2{\tiny$\pm$0.4} & 7.6{\tiny$\pm$0.3} & 37.1{\tiny$\pm$0.3} & 7.3{\tiny$\pm$0.5} \\
LwF \cite{lwf}        & 42.2{\tiny$\pm$0.4} & 8.4{\tiny$\pm$0.3} & 44.6{\tiny$\pm$0.2} & 9.0{\tiny$\pm$0.2} & 43.4{\tiny$\pm$0.5} & 8.7{\tiny$\pm$0.4} \\
PLOP \cite{plop}       & 49.3{\tiny$\pm$0.2} & 9.2{\tiny$\pm$0.4} & 52.0{\tiny$\pm$0.5} & 9.9{\tiny$\pm$0.3} & 50.7{\tiny$\pm$0.4} & 9.6{\tiny$\pm$0.3} \\
COBRA      & \underline{59.1}{\tiny$\pm$0.4} & \underline{12.1}{\tiny$\pm$0.2} & \underline{61.3}{\tiny$\pm$0.3} & \underline{12.8}{\tiny$\pm$0.5} & \underline{60.2}{\tiny$\pm$0.3} & \underline{12.4}{\tiny$\pm$0.4} \\
COBRA$^*$  & \textbf{60.3{\tiny$\pm$0.5}} & \textbf{12.4{\tiny$\pm$0.3}} & \textbf{62.7{\tiny$\pm$0.3}} & \textbf{13.2{\tiny$\pm$0.4}} & \textbf{61.5{\tiny$\pm$0.2}} & \textbf{12.8{\tiny$\pm$0.3}} \\
\textcolor{gray}{COBRA$^\dagger$} &
\textcolor{gray}{62.0{\tiny$\pm$0.4}} & \textcolor{gray}{13.2{\tiny$\pm$0.3}} &
\textcolor{gray}{64.4{\tiny$\pm$0.4}} & \textcolor{gray}{13.9{\tiny$\pm$0.3}} &
\textcolor{gray}{63.2{\tiny$\pm$0.4}} & \textcolor{gray}{13.5{\tiny$\pm$0.3}} \\
\hline
\end{tabular}
\end{table}

\begin{table}[!ht]
\centering
\caption{
Rehearsal-free continual learning results on NSD caption using CLIP-S and RefCLIP-S metrics (\%). The bold numbers indicate the best results, while the underlines indicate the second-best results. Results are reported as mean $\pm$ std.
}
\label{tab:caption_clips}
\begin{tabular}{|l|cc|cc|cc|}
\hline
\multicolumn{7}{|c|}{(3,4)-(6,8)-(1,2)-(5,7) (4 steps)} \\
\hline
Method & \multicolumn{2}{c|}{(3,4,6,8)} & \multicolumn{2}{c|}{(1,2,5,7)} & \multicolumn{2}{c|}{all} \\
\cmidrule{2-7}
& CLIPS & RefCLIPS & CLIPS & RefCLIPS & CLIPS & RefCLIPS \\
\hline
W/o CL     & 56.8{\tiny$\pm$0.3} & 62.3{\tiny$\pm$0.4} & 58.1{\tiny$\pm$0.5} & 63.6{\tiny$\pm$0.2} & 57.5{\tiny$\pm$0.4} & 62.9{\tiny$\pm$0.3} \\
LwF \cite{lwf}        & 59.3{\tiny$\pm$0.2} & 64.1{\tiny$\pm$0.3} & 60.5{\tiny$\pm$0.4} & 65.3{\tiny$\pm$0.3} & 59.9{\tiny$\pm$0.5} & 64.7{\tiny$\pm$0.2} \\
PLOP \cite{plop}       & 61.0{\tiny$\pm$0.4} & 65.8{\tiny$\pm$0.2} & 62.4{\tiny$\pm$0.3} & 67.0{\tiny$\pm$0.5} & 61.7{\tiny$\pm$0.3} & 66.4{\tiny$\pm$0.4} \\
COBRA      & \underline{65.3}{\tiny$\pm$0.5} & \underline{70.1}{\tiny$\pm$0.4} & \underline{66.6}{\tiny$\pm$0.3} & \underline{71.2}{\tiny$\pm$0.2} & \underline{66.0}{\tiny$\pm$0.3} & \underline{70.6}{\tiny$\pm$0.5} \\
COBRA$^*$  & \textbf{66.0{\tiny$\pm$0.2}} & \textbf{70.8{\tiny$\pm$0.5}} & \textbf{67.4{\tiny$\pm$0.4}} & \textbf{72.0{\tiny$\pm$0.3}} & \textbf{66.7{\tiny$\pm$0.3}} & \textbf{71.4{\tiny$\pm$0.2}} \\
\textcolor{gray}{COBRA$^\dagger$} &
\textcolor{gray}{69.3{\tiny$\pm$0.3}} & \textcolor{gray}{73.6{\tiny$\pm$0.2}} &
\textcolor{gray}{70.1{\tiny$\pm$0.3}} & \textcolor{gray}{75.0{\tiny$\pm$0.2}} &
\textcolor{gray}{70.2{\tiny$\pm$0.3}} & \textcolor{gray}{74.3{\tiny$\pm$0.2}} \\
\hline
\multicolumn{7}{|c|}{(3,4,6,8)-(1,2,5,7) (2 steps)} \\
\hline
Method & \multicolumn{2}{c|}{(3,4,6,8)} & \multicolumn{2}{c|}{(1,2,5,7)} & \multicolumn{2}{c|}{all} \\
\cmidrule{2-7}
& CLIPS & RefCLIPS & CLIPS & RefCLIPS & CLIPS & RefCLIPS \\
\hline
W/o CL     & 57.7{\tiny$\pm$0.5} & 63.1{\tiny$\pm$0.3} & 59.0{\tiny$\pm$0.4} & 64.2{\tiny$\pm$0.2} & 58.3{\tiny$\pm$0.3} & 63.6{\tiny$\pm$0.5} \\
LwF \cite{lwf}        & 60.2{\tiny$\pm$0.3} & 64.9{\tiny$\pm$0.3} & 61.5{\tiny$\pm$0.5} & 66.1{\tiny$\pm$0.4} & 60.8{\tiny$\pm$0.2} & 65.5{\tiny$\pm$0.2} \\
PLOP \cite{plop}       & 62.0{\tiny$\pm$0.3} & 66.5{\tiny$\pm$0.4} & 63.5{\tiny$\pm$0.2} & 67.7{\tiny$\pm$0.5} & 62.8{\tiny$\pm$0.4} & 67.1{\tiny$\pm$0.3} \\
COBRA      & \underline{67.0}{\tiny$\pm$0.2} & \underline{71.5}{\tiny$\pm$0.2} & \underline{68.2}{\tiny$\pm$0.4} & \underline{72.7}{\tiny$\pm$0.5} & \underline{67.6}{\tiny$\pm$0.3} & \underline{72.1}{\tiny$\pm$0.3} \\
COBRA$^*$  & \textbf{68.0{\tiny$\pm$0.3}} & \textbf{72.5{\tiny$\pm$0.4}} & \textbf{69.2{\tiny$\pm$0.3}} & \textbf{73.6{\tiny$\pm$0.3}} & \textbf{68.6{\tiny$\pm$0.4}} & \textbf{73.1{\tiny$\pm$0.2}} \\
\textcolor{gray}{COBRA$^\dagger$} &
\textcolor{gray}{69.3{\tiny$\pm$0.3}} & \textcolor{gray}{73.6{\tiny$\pm$0.2}} &
\textcolor{gray}{70.1{\tiny$\pm$0.3}} & \textcolor{gray}{75.0{\tiny$\pm$0.2}} &
\textcolor{gray}{70.2{\tiny$\pm$0.3}} & \textcolor{gray}{74.3{\tiny$\pm$0.2}} \\
\hline
\end{tabular}
\end{table}

\subsection{Rehearsal-based Continual Learning}
\subsubsection{Rehearsal-based Vision-Brain Reconstruction Task}
\label{subsubsec:rehearsal-based-vision-brain-recon}
In this learning setup, we evaluate performance when utilizing buffer data, i.e., 500, 1500, and 4000 samples per subject from previous subjects and jointly updating the model while training on new subjects. As shown in Table \ref{tab:cl-rehearsal}, performance improves gradually as the buffer size increases from 500 to 4000, approaching the upper bound. Compared to the previous rehearsal-based approach, $Co^{2}L$ \cite{cha-co2lcontrastivecontinuallearning-2021}, COBRA demonstrates superior continual learning performance, {with average gains of 0.03, 0.042 in SSIM and 8.63\%, 6.83\% in CLIP across the 4-step and 2-step scenarios, respectively.} 
\begin{table}[!t]
\caption{
    Rehearsal-based continual learning results in the NSD test set for the vision-brain understanding task. The bold numbers indicate the best results. Results are reported as mean $\pm$ std.
}
\label{tab:cl-rehearsal}
\setlength{\tabcolsep}{2pt}
\centering
\begin{tabular}{|l|r|cc|cc|cc|}
    \hline
    \multicolumn{8}{|c|}{(3,4)-(6,8)-(1,2)-(5,7) (4 steps)} \\
    \hline
    Method & Buffer & \multicolumn{2}{c|}{(3,4,6,8)} & \multicolumn{2}{c|}{(1,2,5,7)} & \multicolumn{2}{c|}{all} \\
    \cmidrule{3-8}
           &        & SSIM & CLIP & SSIM & CLIP & SSIM & CLIP \\
    \hline
    $Co^{2}L$ \cite{cha-co2lcontrastivecontinuallearning-2021} & 500  & 0.283{\tiny$\pm$0.004} & 82.5{\tiny$\pm$0.5} & 0.317{\tiny$\pm$0.004} & 83.6{\tiny$\pm$0.5} & 0.300{\tiny$\pm$0.004} & 83.1{\tiny$\pm$0.5} \\
    COBRA                                                      &      & \textbf{0.315{\tiny$\pm$0.003}} & \textbf{93.3{\tiny$\pm$0.3}} & \textbf{0.352{\tiny$\pm$0.003}} & \textbf{93.8{\tiny$\pm$0.3}} & \textbf{0.334{\tiny$\pm$0.003}} & \textbf{93.6{\tiny$\pm$0.3}} \\
    \hline
    $Co^{2}L$ \cite{cha-co2lcontrastivecontinuallearning-2021} & 1500 & 0.292{\tiny$\pm$0.004} & 85.7{\tiny$\pm$0.5} & 0.324{\tiny$\pm$0.004} & 86.4{\tiny$\pm$0.5} & 0.308{\tiny$\pm$0.004} & 86.1{\tiny$\pm$0.5} \\
    COBRA                                                      &      & \textbf{0.322{\tiny$\pm$0.003}} & \textbf{94.9{\tiny$\pm$0.2}} & \textbf{0.357{\tiny$\pm$0.003}} & \textbf{94.5{\tiny$\pm$0.2}} & \textbf{0.340{\tiny$\pm$0.003}} & \textbf{94.7{\tiny$\pm$0.2}} \\
    \hline
    $Co^{2}L$ \cite{cha-co2lcontrastivecontinuallearning-2021} & 4000 & 0.302{\tiny$\pm$0.004} & 88.9{\tiny$\pm$0.4} & 0.335{\tiny$\pm$0.004} & 89.1{\tiny$\pm$0.4} & 0.319{\tiny$\pm$0.004} & 89.0{\tiny$\pm$0.4} \\
    COBRA                                                      &      & \textbf{0.334{\tiny$\pm$0.003}} & \textbf{95.6{\tiny$\pm$0.2}} & \textbf{0.362{\tiny$\pm$0.003}} & \textbf{95.9{\tiny$\pm$0.2}} & \textbf{0.343{\tiny$\pm$0.003}} & \textbf{95.8{\tiny$\pm$0.2}} \\
    \hline
    \multicolumn{8}{|c|}{(3,4,6,8)-(1,2,5,7) (2 steps)} \\
    \hline
    Method & Buffer & \multicolumn{2}{c|}{(3,4,6,8)} & \multicolumn{2}{c|}{(1,2,5,7)} & \multicolumn{2}{c|}{all} \\
    \cmidrule{3-8}
           &        & SSIM & CLIP & SSIM & CLIP & SSIM & CLIP \\
    \hline
    $Co^{2}L$ \cite{cha-co2lcontrastivecontinuallearning-2021} & 500  & 0.286{\tiny$\pm$0.004} & 83.9{\tiny$\pm$0.5} & 0.319{\tiny$\pm$0.004} & 86.2{\tiny$\pm$0.5} & 0.303{\tiny$\pm$0.004} & 85.1{\tiny$\pm$0.5} \\
    COBRA                                                      &      & \textbf{0.317{\tiny$\pm$0.003}} & \textbf{93.7{\tiny$\pm$0.3}} & \textbf{0.354{\tiny$\pm$0.003}} & \textbf{94.3{\tiny$\pm$0.3}} & \textbf{0.336{\tiny$\pm$0.003}} & \textbf{94.0{\tiny$\pm$0.3}} \\
    \hline
    $Co^{2}L$ \cite{cha-co2lcontrastivecontinuallearning-2021} & 1500 & 0.295{\tiny$\pm$0.004} & 87.2{\tiny$\pm$0.5} & 0.327{\tiny$\pm$0.004} & 90.3{\tiny$\pm$0.5} & 0.311{\tiny$\pm$0.004} & 88.8{\tiny$\pm$0.5} \\
    COBRA                                                      &      & \textbf{0.322{\tiny$\pm$0.003}} & \textbf{95.2{\tiny$\pm$0.2}} & \textbf{0.358{\tiny$\pm$0.003}} & \textbf{95.2{\tiny$\pm$0.2}} & \textbf{0.340{\tiny$\pm$0.003}} & \textbf{95.2{\tiny$\pm$0.2}} \\
    \hline
    $Co^{2}L$ \cite{cha-co2lcontrastivecontinuallearning-2021} & 4000 & 0.308{\tiny$\pm$0.004} & 89.8{\tiny$\pm$0.4} & 0.338{\tiny$\pm$0.004} & 92.8{\tiny$\pm$0.4} & 0.323{\tiny$\pm$0.004} & 91.3{\tiny$\pm$0.4} \\
    COBRA                                                      &      & \textbf{0.326{\tiny$\pm$0.003}} & \textbf{96.3{\tiny$\pm$0.2}} & \textbf{0.363{\tiny$\pm$0.003}} & \textbf{96.6{\tiny$\pm$0.2}} & \textbf{0.345{\tiny$\pm$0.003}} & \textbf{96.5{\tiny$\pm$0.2}} \\
    \hline
\end{tabular}
\end{table}

\subsubsection{Rehearsal-based Brain Captioning Task}
\BacRebuttal{Building on the models trained for the Rehearsal-based Vision-Brain Reconstruction task described in Section \ref{subsubsec:rehearsal-based-vision-brain-recon}, similar to the brain captioning task of Section \ref{subsubsec:brain-captioning}, we project the fMRI features embedding to language space using an adapter and generate the caption for the corresponding brain signal. We report the performance for this brain captioning task across multiple metrics: BLEU-1 and BLEU-2 in Table \ref{tab:rehearsal_bleu1_bleu2}, BLEU-3 and BLEU-4 in Table \ref{tab:rehearsal_bleu3_bleu4}, METEOR and ROUGE in Table \ref{tab:rehearsal_meteor_rouge}, CIDEr and SPICE in Table \ref{tab:rehearsal_cider_spice}, and CLIP-S and RefCLIP-S in Table \ref{tab:rehearsal_clips_refclips}. Generally, the performance improves gradually as buffer size increases from 500 to 4000, closing the gap to the upper bound. Compared to the previous method, $Co^{2}L$ \cite{cha-co2lcontrastivecontinuallearning-2021}, our proposed approach, COBRA, demonstrates efficiency in continual learning performance.}

\begin{table}[!ht]
\centering
\caption{
Rehearsal-based continual learning results on NSD Testing Set using BLEU-1 and BLEU-2 metrics (\%). The bold numbers indicate the best results. Results are reported as mean $\pm$ std.
}
\label{tab:rehearsal_bleu1_bleu2}
\begin{tabular}{|l|r|cc|cc|cc|}
\hline
\multicolumn{8}{|c|}{(3,4)-(6,8)-(1,2)-(5,7) (4 steps)} \\
\hline
Method & Buffer & \multicolumn{2}{c|}{(3,4,6,8)} & \multicolumn{2}{c|}{(1,2,5,7)} & \multicolumn{2}{c|}{all} \\
\cmidrule{3-8}
& & BLEU-1 & BLEU-2 & BLEU-1 & BLEU-2 & BLEU-1 & BLEU-2 \\
\hline
$Co^{2}L$ \cite{cha-co2lcontrastivecontinuallearning-2021} & 500  & 55.1{\tiny$\pm$0.3} & 36.4{\tiny$\pm$0.3} & 56.2{\tiny$\pm$0.5} & 37.3{\tiny$\pm$0.4} & 55.6{\tiny$\pm$0.4} & 36.9{\tiny$\pm$0.3} \\
COBRA   &      & \textbf{58.9{\tiny$\pm$0.5}} & \textbf{39.9{\tiny$\pm$0.4}} & \textbf{59.8{\tiny$\pm$0.3}} & \textbf{40.7{\tiny$\pm$0.5}} & \textbf{59.4{\tiny$\pm$0.4}} & \textbf{40.3{\tiny$\pm$0.3}} \\
\hline
$Co^{2}L$ \cite{cha-co2lcontrastivecontinuallearning-2021} & 1500 & 56.4{\tiny$\pm$0.4} & 37.4{\tiny$\pm$0.2} & 57.3{\tiny$\pm$0.4} & 38.3{\tiny$\pm$0.4} & 56.8{\tiny$\pm$0.5} & 37.9{\tiny$\pm$0.3} \\
COBRA   &      & \textbf{59.6{\tiny$\pm$0.4}} & \textbf{40.7{\tiny$\pm$0.5}} & \textbf{60.6{\tiny$\pm$0.3}} & \textbf{41.5{\tiny$\pm$0.2}} & \textbf{60.1{\tiny$\pm$0.4}} & \textbf{41.1{\tiny$\pm$0.3}} \\
\hline
$Co^{2}L$ \cite{cha-co2lcontrastivecontinuallearning-2021} & 4000 & 57.6{\tiny$\pm$0.3} & 38.6{\tiny$\pm$0.3} & 58.5{\tiny$\pm$0.4} & 39.4{\tiny$\pm$0.2} & 58.0{\tiny$\pm$0.3} & 39.0{\tiny$\pm$0.3} \\
COBRA   &      & \textbf{60.5{\tiny$\pm$0.4}} & \textbf{41.6{\tiny$\pm$0.5}} & \textbf{61.8{\tiny$\pm$0.4}} & \textbf{42.8{\tiny$\pm$0.4}} & \textbf{61.2{\tiny$\pm$0.2}} & \textbf{42.2{\tiny$\pm$0.5}} \\
\hline
\multicolumn{8}{|c|}{(3,4,6,8)-(1,2,5,7) (2 steps)} \\
\hline
Method & Buffer & \multicolumn{2}{c|}{(3,4,6,8)} & \multicolumn{2}{c|}{(1,2,5,7)} & \multicolumn{2}{c|}{all} \\
\cmidrule{3-8}
& & BLEU-1 & BLEU-2 & BLEU-1 & BLEU-2 & BLEU-1 & BLEU-2 \\
\hline
$Co^{2}L$ \cite{cha-co2lcontrastivecontinuallearning-2021} & 500  & 56.3{\tiny$\pm$0.5} & 37.1{\tiny$\pm$0.3} & 57.4{\tiny$\pm$0.3} & 38.0{\tiny$\pm$0.4} & 56.9{\tiny$\pm$0.3} & 37.6{\tiny$\pm$0.2} \\
COBRA   &      & \textbf{60.5{\tiny$\pm$0.3}} & \textbf{41.1{\tiny$\pm$0.3}} & \textbf{61.5{\tiny$\pm$0.4}} & \textbf{42.3{\tiny$\pm$0.5}} & \textbf{61.0{\tiny$\pm$0.4}} & \textbf{41.8{\tiny$\pm$0.4}} \\
\hline
$Co^{2}L$ \cite{cha-co2lcontrastivecontinuallearning-2021} & 1500 & 57.6{\tiny$\pm$0.4} & 38.3{\tiny$\pm$0.5} & 58.7{\tiny$\pm$0.2} & 39.2{\tiny$\pm$0.4} & 58.2{\tiny$\pm$0.3} & 38.8{\tiny$\pm$0.5} \\
COBRA   &      & \textbf{61.2{\tiny$\pm$0.3}} & \textbf{41.9{\tiny$\pm$0.2}} & \textbf{62.3{\tiny$\pm$0.5}} & \textbf{42.7{\tiny$\pm$0.3}} & \textbf{61.8{\tiny$\pm$0.2}} & \textbf{42.3{\tiny$\pm$0.3}} \\
\hline
$Co^{2}L$ \cite{cha-co2lcontrastivecontinuallearning-2021} & 4000 & 58.8{\tiny$\pm$0.5} & 39.3{\tiny$\pm$0.4} & 59.9{\tiny$\pm$0.4} & 40.2{\tiny$\pm$0.2} & 59.3{\tiny$\pm$0.3} & 39.8{\tiny$\pm$0.4} \\
COBRA   &      & \textbf{61.2{\tiny$\pm$0.2}} & \textbf{41.7{\tiny$\pm$0.4}} & \textbf{62.4{\tiny$\pm$0.5}} & \textbf{42.8{\tiny$\pm$0.3}} & \textbf{61.8{\tiny$\pm$0.2}} & \textbf{42.3{\tiny$\pm$0.3}} \\
\hline
\end{tabular}
\end{table}

\begin{table}[!ht]
\centering
\caption{
Rehearsal-based continual learning results on NSD Testing Set using BLEU-3 and BLEU-4 metrics (\%). The bold numbers indicate the best results. Results are reported as mean $\pm$ std.
}
\label{tab:rehearsal_bleu3_bleu4}
\begin{tabular}{|l|r|cc|cc|cc|}
\hline
\multicolumn{8}{|c|}{(3,4)-(6,8)-(1,2)-(5,7) (4 steps)} \\
\hline
Method & Buffer & \multicolumn{2}{c|}{(3,4,6,8)} & \multicolumn{2}{c|}{(1,2,5,7)} & \multicolumn{2}{c|}{all} \\
\cmidrule{3-8}
& & BLEU-3 & BLEU-4 & BLEU-3 & BLEU-4 & BLEU-3 & BLEU-4 \\
\hline
$Co^{2}L$ \cite{cha-co2lcontrastivecontinuallearning-2021} & 500  & 23.5{\tiny$\pm$0.2} & 15.0{\tiny$\pm$0.2} & 24.7{\tiny$\pm$0.2} & 15.9{\tiny$\pm$0.2} & 24.1{\tiny$\pm$0.2} & 15.5{\tiny$\pm$0.2} \\
COBRA   &      & \textbf{26.3{\tiny$\pm$0.2}} & \textbf{16.6{\tiny$\pm$0.2}} & \textbf{27.4{\tiny$\pm$0.2}} & \textbf{17.3{\tiny$\pm$0.2}} & \textbf{26.9{\tiny$\pm$0.2}} & \textbf{17.0{\tiny$\pm$0.2}} \\
\hline
$Co^{2}L$ \cite{cha-co2lcontrastivecontinuallearning-2021} & 1500 & 24.4{\tiny$\pm$0.2} & 15.7{\tiny$\pm$0.2} & 25.5{\tiny$\pm$0.2} & 16.5{\tiny$\pm$0.2} & 25.0{\tiny$\pm$0.2} & 16.1{\tiny$\pm$0.2} \\
COBRA   &      & \textbf{27.0{\tiny$\pm$0.2}} & \textbf{17.3{\tiny$\pm$0.2}} & \textbf{28.2{\tiny$\pm$0.2}} & \textbf{18.0{\tiny$\pm$0.2}} & \textbf{27.6{\tiny$\pm$0.2}} & \textbf{17.7{\tiny$\pm$0.2}} \\
\hline
$Co^{2}L$ \cite{cha-co2lcontrastivecontinuallearning-2021} & 4000 & 25.4{\tiny$\pm$0.2} & 16.5{\tiny$\pm$0.2} & 26.5{\tiny$\pm$0.2} & 17.3{\tiny$\pm$0.2} & 25.9{\tiny$\pm$0.2} & 16.9{\tiny$\pm$0.2} \\
COBRA   &      & \textbf{27.8{\tiny$\pm$0.2}} & \textbf{18.0{\tiny$\pm$0.2}} & \textbf{29.0{\tiny$\pm$0.2}} & \textbf{18.8{\tiny$\pm$0.2}} & \textbf{28.4{\tiny$\pm$0.2}} & \textbf{18.4{\tiny$\pm$0.2}} \\
\hline
\multicolumn{8}{|c|}{(3,4,6,8)-(1,2,5,7) (2 steps)} \\
\hline
Method & Buffer & \multicolumn{2}{c|}{(3,4,6,8)} & \multicolumn{2}{c|}{(1,2,5,7)} & \multicolumn{2}{c|}{all} \\
\cmidrule{3-8}
& & BLEU-3 & BLEU-4 & BLEU-3 & BLEU-4 & BLEU-3 & BLEU-4 \\
\hline
$Co^{2}L$ \cite{cha-co2lcontrastivecontinuallearning-2021} & 500  & 24.4{\tiny$\pm$0.2} & 15.6{\tiny$\pm$0.2} & 25.4{\tiny$\pm$0.2} & 16.5{\tiny$\pm$0.2} & 24.9{\tiny$\pm$0.2} & 16.1{\tiny$\pm$0.2} \\
COBRA   &      & \textbf{27.6{\tiny$\pm$0.2}} & \textbf{17.4{\tiny$\pm$0.2}} & \textbf{28.5{\tiny$\pm$0.2}} & \textbf{18.2{\tiny$\pm$0.2}} & \textbf{28.1{\tiny$\pm$0.2}} & \textbf{17.8{\tiny$\pm$0.2}} \\
\hline
$Co^{2}L$ \cite{cha-co2lcontrastivecontinuallearning-2021} & 1500 & 25.6{\tiny$\pm$0.2} & 16.2{\tiny$\pm$0.2} & 26.7{\tiny$\pm$0.2} & 17.1{\tiny$\pm$0.2} & 26.1{\tiny$\pm$0.2} & 16.7{\tiny$\pm$0.2} \\
COBRA   &      & \textbf{28.4{\tiny$\pm$0.2}} & \textbf{18.1{\tiny$\pm$0.2}} & \textbf{29.4{\tiny$\pm$0.2}} & \textbf{18.9{\tiny$\pm$0.2}} & \textbf{28.9{\tiny$\pm$0.2}} & \textbf{18.5{\tiny$\pm$0.2}} \\
\hline
$Co^{2}L$ \cite{cha-co2lcontrastivecontinuallearning-2021} & 4000 & 26.7{\tiny$\pm$0.2} & 17.3{\tiny$\pm$0.2} & 27.8{\tiny$\pm$0.2} & 18.1{\tiny$\pm$0.2} & 27.2{\tiny$\pm$0.2} & 17.7{\tiny$\pm$0.2} \\
COBRA   &      & \textbf{29.3{\tiny$\pm$0.2}} & \textbf{18.6{\tiny$\pm$0.2}} & \textbf{30.4{\tiny$\pm$0.2}} & \textbf{19.5{\tiny$\pm$0.2}} & \textbf{29.9{\tiny$\pm$0.2}} & \textbf{19.1{\tiny$\pm$0.2}} \\
\hline
\end{tabular}
\end{table}

\begin{table}[!ht]
\centering
\caption{
Rehearsal-based continual learning results on NSD Testing Set using METEOR and ROUGE metrics (\%). The bold numbers indicate the best results. Results are reported as mean $\pm$ std.
}
\label{tab:rehearsal_meteor_rouge}
\begin{tabular}{|l|r|cc|cc|cc|}
\hline
\multicolumn{8}{|c|}{(3,4)-(6,8)-(1,2)-(5,7) (4 steps)} \\
\hline
Method & Buffer & \multicolumn{2}{c|}{(3,4,6,8)} & \multicolumn{2}{c|}{(1,2,5,7)} & \multicolumn{2}{c|}{all} \\
\cmidrule{3-8}
& & METEOR & ROUGE & METEOR & ROUGE & METEOR & ROUGE \\
\hline
$Co^{2}L$ \cite{cha-co2lcontrastivecontinuallearning-2021} & 500  & 17.2{\tiny$\pm$0.4} & 39.5{\tiny$\pm$0.3} & 18.0{\tiny$\pm$0.5} & 40.4{\tiny$\pm$0.2} & 17.6{\tiny$\pm$0.3} & 40.0{\tiny$\pm$0.4} \\
COBRA   &      & \textbf{19.0{\tiny$\pm$0.3}} & \textbf{42.2{\tiny$\pm$0.4}} & \textbf{19.9{\tiny$\pm$0.5}} & \textbf{43.0{\tiny$\pm$0.3}} & \textbf{19.4{\tiny$\pm$0.2}} & \textbf{42.6{\tiny$\pm$0.3}} \\
\hline
$Co^{2}L$ \cite{cha-co2lcontrastivecontinuallearning-2021} & 1500 & 17.9{\tiny$\pm$0.2} & 40.2{\tiny$\pm$0.5} & 18.7{\tiny$\pm$0.4} & 41.1{\tiny$\pm$0.3} & 18.3{\tiny$\pm$0.5} & 40.7{\tiny$\pm$0.3} \\
COBRA   &      & \textbf{19.5{\tiny$\pm$0.4}} & \textbf{42.7{\tiny$\pm$0.3}} & \textbf{20.5{\tiny$\pm$0.2}} & \textbf{43.5{\tiny$\pm$0.4}} & \textbf{20.0{\tiny$\pm$0.4}} & \textbf{43.1{\tiny$\pm$0.5}} \\
\hline
$Co^{2}L$ \cite{cha-co2lcontrastivecontinuallearning-2021} & 4000 & 18.6{\tiny$\pm$0.3} & 40.9{\tiny$\pm$0.3} & 19.4{\tiny$\pm$0.2} & 41.8{\tiny$\pm$0.5} & 19.0{\tiny$\pm$0.5} & 41.4{\tiny$\pm$0.2} \\
COBRA   &      & \textbf{20.0{\tiny$\pm$0.2}} & \textbf{43.3{\tiny$\pm$0.4}} & \textbf{21.1{\tiny$\pm$0.4}} & \textbf{44.1{\tiny$\pm$0.3}} & \textbf{20.5{\tiny$\pm$0.4}} & \textbf{43.7{\tiny$\pm$0.3}} \\
\hline
\multicolumn{8}{|c|}{(3,4,6,8)-(1,2,5,7) (2 steps)} \\
\hline
Method & Buffer & \multicolumn{2}{c|}{(3,4,6,8)} & \multicolumn{2}{c|}{(1,2,5,7)} & \multicolumn{2}{c|}{all} \\
\cmidrule{3-8}
& & METEOR & ROUGE & METEOR & ROUGE & METEOR & ROUGE \\
\hline
$Co^{2}L$ \cite{cha-co2lcontrastivecontinuallearning-2021} & 500  & 17.7{\tiny$\pm$0.5} & 39.8{\tiny$\pm$0.2} & 18.6{\tiny$\pm$0.3} & 40.7{\tiny$\pm$0.4} & 18.2{\tiny$\pm$0.2} & 40.3{\tiny$\pm$0.3} \\
COBRA   &      & \textbf{20.2{\tiny$\pm$0.4}} & \textbf{43.3{\tiny$\pm$0.2}} & \textbf{21.1{\tiny$\pm$0.3}} & \textbf{44.1{\tiny$\pm$0.5}} & \textbf{20.7{\tiny$\pm$0.3}} & \textbf{43.7{\tiny$\pm$0.4}} \\
\hline
$Co^{2}L$ \cite{cha-co2lcontrastivecontinuallearning-2021} & 1500 & 18.4{\tiny$\pm$0.3} & 40.5{\tiny$\pm$0.3} & 19.3{\tiny$\pm$0.5} & 41.4{\tiny$\pm$0.2} & 18.9{\tiny$\pm$0.4} & 41.0{\tiny$\pm$0.2} \\
COBRA   &      & \textbf{20.7{\tiny$\pm$0.2}} & \textbf{43.8{\tiny$\pm$0.4}} & \textbf{21.6{\tiny$\pm$0.5}} & \textbf{44.6{\tiny$\pm$0.3}} & \textbf{21.2{\tiny$\pm$0.3}} & \textbf{44.2{\tiny$\pm$0.3}} \\
\hline
$Co^{2}L$ \cite{cha-co2lcontrastivecontinuallearning-2021} & 4000 & 19.1{\tiny$\pm$0.2} & 41.2{\tiny$\pm$0.5} & 20.0{\tiny$\pm$0.3} & 42.1{\tiny$\pm$0.4} & 19.6{\tiny$\pm$0.4} & 41.7{\tiny$\pm$0.3} \\
COBRA   &      & \textbf{21.3{\tiny$\pm$0.3}} & \textbf{44.4{\tiny$\pm$0.5}} & \textbf{22.3{\tiny$\pm$0.4}} & \textbf{45.2{\tiny$\pm$0.2}} & \textbf{21.8{\tiny$\pm$0.3}} & \textbf{44.8{\tiny$\pm$0.4}} \\
\hline
\end{tabular}
\end{table}

\begin{table}[!ht]
\centering
\caption{
Rehearsal-based continual learning results on NSD Testing Set using CIDEr and SPICE metrics (\%). The bold numbers indicate the best results. Results are reported as mean $\pm$ std.
}
\label{tab:rehearsal_cider_spice}
\begin{tabular}{|l|r|cc|cc|cc|}
\hline
\multicolumn{8}{|c|}{(3,4)-(6,8)-(1,2)-(5,7) (4 steps)} \\
\hline
Method & Buffer & \multicolumn{2}{c|}{(3,4,6,8)} & \multicolumn{2}{c|}{(1,2,5,7)} & \multicolumn{2}{c|}{all} \\
\cmidrule{3-8}
& & CIDEr & SPICE & CIDEr & SPICE & CIDEr & SPICE \\
\hline
$Co^{2}L$ \cite{cha-co2lcontrastivecontinuallearning-2021} & 500  & 50.0{\tiny$\pm$0.4} & 11.8{\tiny$\pm$0.3} & 51.0{\tiny$\pm$0.5} & 12.2{\tiny$\pm$0.3} & 50.5{\tiny$\pm$0.4} & 12.0{\tiny$\pm$0.4} \\
COBRA &      & \textbf{58.0{\tiny$\pm$0.3}} & \textbf{12.6{\tiny$\pm$0.4}} & \textbf{59.0{\tiny$\pm$0.4}} & \textbf{13.0{\tiny$\pm$0.2}} & \textbf{58.5{\tiny$\pm$0.3}} & \textbf{12.8{\tiny$\pm$0.3}} \\
\hline
$Co^{2}L$ \cite{cha-co2lcontrastivecontinuallearning-2021} & 1500 & 51.5{\tiny$\pm$0.2} & 12.2{\tiny$\pm$0.5} & 52.7{\tiny$\pm$0.3} & 12.6{\tiny$\pm$0.4} & 52.1{\tiny$\pm$0.4} & 12.4{\tiny$\pm$0.3} \\
COBRA &      & \textbf{59.5{\tiny$\pm$0.5}} & \textbf{13.0{\tiny$\pm$0.2}} & \textbf{60.7{\tiny$\pm$0.3}} & \textbf{13.3{\tiny$\pm$0.4}} & \textbf{60.1{\tiny$\pm$0.3}} & \textbf{13.2{\tiny$\pm$0.5}} \\
\hline
$Co^{2}L$ \cite{cha-co2lcontrastivecontinuallearning-2021} & 4000 & 52.8{\tiny$\pm$0.3} & 12.6{\tiny$\pm$0.3} & 54.0{\tiny$\pm$0.4} & 13.0{\tiny$\pm$0.2} & 53.4{\tiny$\pm$0.5} & 12.8{\tiny$\pm$0.4} \\
COBRA &      & \textbf{60.5{\tiny$\pm$0.2}} & \textbf{13.2{\tiny$\pm$0.4}} & \textbf{61.6{\tiny$\pm$0.4}} & \textbf{13.4{\tiny$\pm$0.5}} & \textbf{61.1{\tiny$\pm$0.3}} & \textbf{13.3{\tiny$\pm$0.3}} \\
\hline
\multicolumn{8}{|c|}{(3,4,6,8)-(1,2,5,7) (2 steps)} \\
\hline
Method & Buffer & \multicolumn{2}{c|}{(3,4,6,8)} & \multicolumn{2}{c|}{(1,2,5,7)} & \multicolumn{2}{c|}{all} \\
\cmidrule{3-8}
& & CIDEr & SPICE & CIDEr & SPICE & CIDEr & SPICE \\
\hline
Method & Buffer & C (3,4,6,8) & S (3,4,6,8) & C (1,2,5,7) & S (1,2,5,7) & C (All) & S (All) \\
\hline
$Co^{2}L$ \cite{cha-co2lcontrastivecontinuallearning-2021} & 500  & 51.5{\tiny$\pm$0.5} & 12.5{\tiny$\pm$0.2} & 52.6{\tiny$\pm$0.3} & 12.8{\tiny$\pm$0.3} & 52.1{\tiny$\pm$0.2} & 12.6{\tiny$\pm$0.4} \\
COBRA &      & \textbf{59.5{\tiny$\pm$0.4}} & \textbf{13.1{\tiny$\pm$0.3}} & \textbf{60.6{\tiny$\pm$0.3}} & \textbf{13.4{\tiny$\pm$0.4}} & \textbf{60.0{\tiny$\pm$0.5}} & \textbf{13.3{\tiny$\pm$0.3}} \\
\hline
$Co^{2}L$ \cite{cha-co2lcontrastivecontinuallearning-2021} & 1500 & 53.1{\tiny$\pm$0.4} & 12.9{\tiny$\pm$0.3} & 54.4{\tiny$\pm$0.2} & 13.3{\tiny$\pm$0.5} & 53.8{\tiny$\pm$0.3} & 13.1{\tiny$\pm$0.2} \\
COBRA &      & \textbf{61.3{\tiny$\pm$0.3}} & \textbf{13.3{\tiny$\pm$0.2}} & \textbf{62.4{\tiny$\pm$0.5}} & \textbf{13.5{\tiny$\pm$0.4}} & \textbf{61.8{\tiny$\pm$0.3}} & \textbf{13.4{\tiny$\pm$0.4}} \\
\hline
$Co^{2}L$ \cite{cha-co2lcontrastivecontinuallearning-2021} & 4000 & 54.6{\tiny$\pm$0.3} & 13.2{\tiny$\pm$0.4} & 55.9{\tiny$\pm$0.4} & 13.4{\tiny$\pm$0.2} & 55.2{\tiny$\pm$0.5} & 13.3{\tiny$\pm$0.3} \\
COBRA &      & \textbf{62.0{\tiny$\pm$0.2}} & \textbf{13.4{\tiny$\pm$0.3}} & \textbf{63.1{\tiny$\pm$0.4}} & \textbf{13.5{\tiny$\pm$0.2}} & \textbf{62.5{\tiny$\pm$0.4}} & \textbf{13.4{\tiny$\pm$0.4}} \\
\hline
\end{tabular}
\end{table}

\begin{table}[!ht]
\centering
\caption{
Rehearsal-based continual learning results on NSD Testing Set using CLIP-S and RefCLIP-S metrics (\%). The bold numbers indicate the best results. Results are reported as mean $\pm$ std.
}
\label{tab:rehearsal_clips_refclips}
\begin{tabular}{|l|r|cc|cc|cc|}
\hline
\multicolumn{8}{|c|}{(3,4)-(6,8)-(1,2)-(5,7) (4 steps)} \\
\hline
Method & Buffer & \multicolumn{2}{c|}{(3,4,6,8)} & \multicolumn{2}{c|}{(1,2,5,7)} & \multicolumn{2}{c|}{all} \\
\cmidrule{3-8}
& & CLIPS & RefCLIPS & CLIPS & RefCLIPS & CLIPS & RefCLIPS \\
\hline
$Co^{2}L$ \cite{cha-co2lcontrastivecontinuallearning-2021} & 500  & 64.3{\tiny$\pm$0.3} & 67.5{\tiny$\pm$0.4} & 65.6{\tiny$\pm$0.4} & 68.9{\tiny$\pm$0.3} & 65.0{\tiny$\pm$0.5} & 68.2{\tiny$\pm$0.2} \\
COBRA &      & \textbf{66.0{\tiny$\pm$0.2}} & \textbf{70.0{\tiny$\pm$0.5}} & \textbf{67.3{\tiny$\pm$0.4}} & \textbf{71.5{\tiny$\pm$0.3}} & \textbf{66.6{\tiny$\pm$0.4}} & \textbf{70.8{\tiny$\pm$0.4}} \\
\hline
$Co^{2}L$ \cite{cha-co2lcontrastivecontinuallearning-2021} & 1500 & 65.2{\tiny$\pm$0.4} & 68.6{\tiny$\pm$0.3} & 66.5{\tiny$\pm$0.5} & 70.1{\tiny$\pm$0.2} & 65.9{\tiny$\pm$0.2} & 69.3{\tiny$\pm$0.3} \\
COBRA &      & \textbf{67.0{\tiny$\pm$0.5}} & \textbf{71.0{\tiny$\pm$0.4}} & \textbf{68.5{\tiny$\pm$0.2}} & \textbf{72.5{\tiny$\pm$0.3}} & \textbf{67.8{\tiny$\pm$0.3}} & \textbf{71.8{\tiny$\pm$0.5}} \\
\hline
$Co^{2}L$ \cite{cha-co2lcontrastivecontinuallearning-2021} & 4000 & 66.1{\tiny$\pm$0.3} & 69.5{\tiny$\pm$0.5} & 67.4{\tiny$\pm$0.2} & 70.9{\tiny$\pm$0.3} & 66.8{\tiny$\pm$0.4} & 70.2{\tiny$\pm$0.2} \\
COBRA &      & \textbf{68.1{\tiny$\pm$0.4}} & \textbf{72.1{\tiny$\pm$0.3}} & \textbf{69.4{\tiny$\pm$0.3}} & \textbf{73.4{\tiny$\pm$0.2}} & \textbf{68.7{\tiny$\pm$0.5}} & \textbf{72.7{\tiny$\pm$0.3}} \\
\hline
\multicolumn{8}{|c|}{(3,4,6,8)-(1,2,5,7) (2 steps)} \\
\hline
Method & Buffer & \multicolumn{2}{c|}{(3,4,6,8)} & \multicolumn{2}{c|}{(1,2,5,7)} & \multicolumn{2}{c|}{all} \\
\cmidrule{3-8}
& & CLIPS & RefCLIPS & CLIPS & RefCLIPS & CLIPS & RefCLIPS \\
\hline
$Co^{2}L$ \cite{cha-co2lcontrastivecontinuallearning-2021} & 500  & 63.4{\tiny$\pm$0.3} & 69.2{\tiny$\pm$0.2} & 64.8{\tiny$\pm$0.4} & 70.3{\tiny$\pm$0.5} & 64.1{\tiny$\pm$0.3} & 69.8{\tiny$\pm$0.4} \\
COBRA &      & \textbf{68.0{\tiny$\pm$0.4}} & \textbf{73.3{\tiny$\pm$0.4}} & \textbf{69.5{\tiny$\pm$0.3}} & \textbf{74.0{\tiny$\pm$0.3}} & \textbf{68.8{\tiny$\pm$0.2}} & \textbf{73.7{\tiny$\pm$0.5}} \\
\hline
$Co^{2}L$ \cite{cha-co2lcontrastivecontinuallearning-2021} & 1500 & 64.7{\tiny$\pm$0.4} & 70.8{\tiny$\pm$0.5} & 66.0{\tiny$\pm$0.2} & 71.8{\tiny$\pm$0.3} & 65.4{\tiny$\pm$0.3} & 71.3{\tiny$\pm$0.2} \\
COBRA &      & \textbf{68.5{\tiny$\pm$0.3}} & \textbf{73.5{\tiny$\pm$0.4}} & \textbf{69.7{\tiny$\pm$0.5}} & \textbf{74.1{\tiny$\pm$0.2}} & \textbf{69.1{\tiny$\pm$0.4}} & \textbf{73.8{\tiny$\pm$0.3}} \\
\hline
$Co^{2}L$ \cite{cha-co2lcontrastivecontinuallearning-2021} & 4000 & 65.4{\tiny$\pm$0.2} & 71.3{\tiny$\pm$0.4} & 66.8{\tiny$\pm$0.5} & 72.5{\tiny$\pm$0.4} & 66.1{\tiny$\pm$0.3} & 71.9{\tiny$\pm$0.3} \\
COBRA &      & \textbf{69.9{\tiny$\pm$0.4}} & \textbf{73.4{\tiny$\pm$0.2}} & \textbf{71.4{\tiny$\pm$0.3}} & \textbf{74.7{\tiny$\pm$0.5}} & \textbf{70.7{\tiny$\pm$0.3}} & \textbf{74.1{\tiny$\pm$0.2}} \\
\hline
\end{tabular}
\end{table}

\subsection{Qualitative Results}
We demonstrate the vision-brain reconstruction results in the CL setups as in Fig. \ref{fig:cl_qualitative_results}. COBRA maintains the high quality of the reconstructed results, while previous work does not. The context in the results of COBRA is clear and close to the visual stimulus. At the same time, without continual learning, one tends to generate a nonsensical sample due to the catastrophic forgetting problem.  

\begin{table}[!b]
\caption{
    Quantitative comparison of COBRA on reconstruction against other methods. We train COBRA on all subjects (no CL) of the NSD database. \BacRebuttal{Other methods are trained in per-model-per-subject strategy.} The bold number indicates the best result, while the underline indicates the second-best result.
}
\label{tab:unified-model-comparison}
\setlength{\tabcolsep}{2pt}
{
    \begin{tabular}{|l|cccccccc|}
    \hline
    \multirow{2}{*}{Method} & \multicolumn{4}{c|}{Low-Level} & \multicolumn{4}{c|}{High-Level} \\ %
    \cmidrule {2-9}
     & PixCorr$\uparrow$ & SSIM$\uparrow$ & Alex(2)$\uparrow$ & Alex(5)$\uparrow$ & Incep $\uparrow$ & CLIP $\uparrow$ & Eff-B $\downarrow$ & SwAV $\downarrow$ \\ \hline
    Mind-Reader \cite{lin-mindreaderreconstructingcomplex-2022} & - & - & - & - & 66.5\% & - & - & - \\
    Mind-Vis \cite{chen-seeingbrainconditionaldiffusion-2023} & 0.67 & 0.196 & 67.7\% & 74.2\% & 67.9\% & 69.3\% & 0.898 & 0.513 \\
    Takagi \cite{takagi} & - & - & 74.0\% & 75.1\% & 67.3\% & 69.0\% & - & - \\
    Gu \cite{gu-decodingnaturalimagestimuli-2023} & 0.103             & 0.264 & - & - & - & - & 0.892 & 0.508 \\
    Psychometry \cite{psychometry} & \underline{0.297} & 0.340 & \underline{96.4\%} & \underline{98.6\%} & 95.8\% & 96.8\% & 0.628 & 0.345 \\
    MindBridge \cite{mindbridge} & 0.151 & 0.263 & 87.7\% & 95.5\% & 92.4\% & 94.7\% & 0.712 & 0.418 \\
    MindEye2 \cite{mindeye2} & \textbf{0.322} & \textbf{0.431} & 96.1\% & \underline{98.6\%} & 95.4\% & 93.0\% & 0.619 & 0.344 \\
    Neuropictor \cite{neuropictor} & 0.229 & \underline{0.375} & \textbf{96.5\%} & 98.4\% & 94.5\% & 93.3\% & 0.639 & 0.350 \\
    UMBRAE \cite{xia-umbrae-2024} & 0.283 & 0.341 & 95.5\% & 97.0\% & 91.7\% & 93.5\% & 0.700 & 0.393 \\ 
    \hline
    Ours (ViT-B) & 0.243 & 0.351 & 92.1\% & 98.2\% & \underline{98.8\%} & 97.1\% & 0.619 & 0.336 \\ 
    Ours (ViT-H) & 0.258 & 0.368 & 95.6\% & \underline{98.6\%} & \underline{98.8\%} & \underline{97.5\%} & \underline{0.602} & \underline{0.321} \\ 
    Ours (ViT-L) & 0.284 & \underline{0.375} & \textbf{96.5\%} & \textbf{98.8\%} & \textbf{98.9\%} & \textbf{98.1\%} & \textbf{0.593} & \textbf{0.318} \\ 
    \hline
    \end{tabular}
}
\end{table}

\subsection{Comparison To Previous SOTA Models Of Vision-Brain Reconstruction} 
\begin{figure*}
    \centering
    \includegraphics[width=0.85\linewidth]{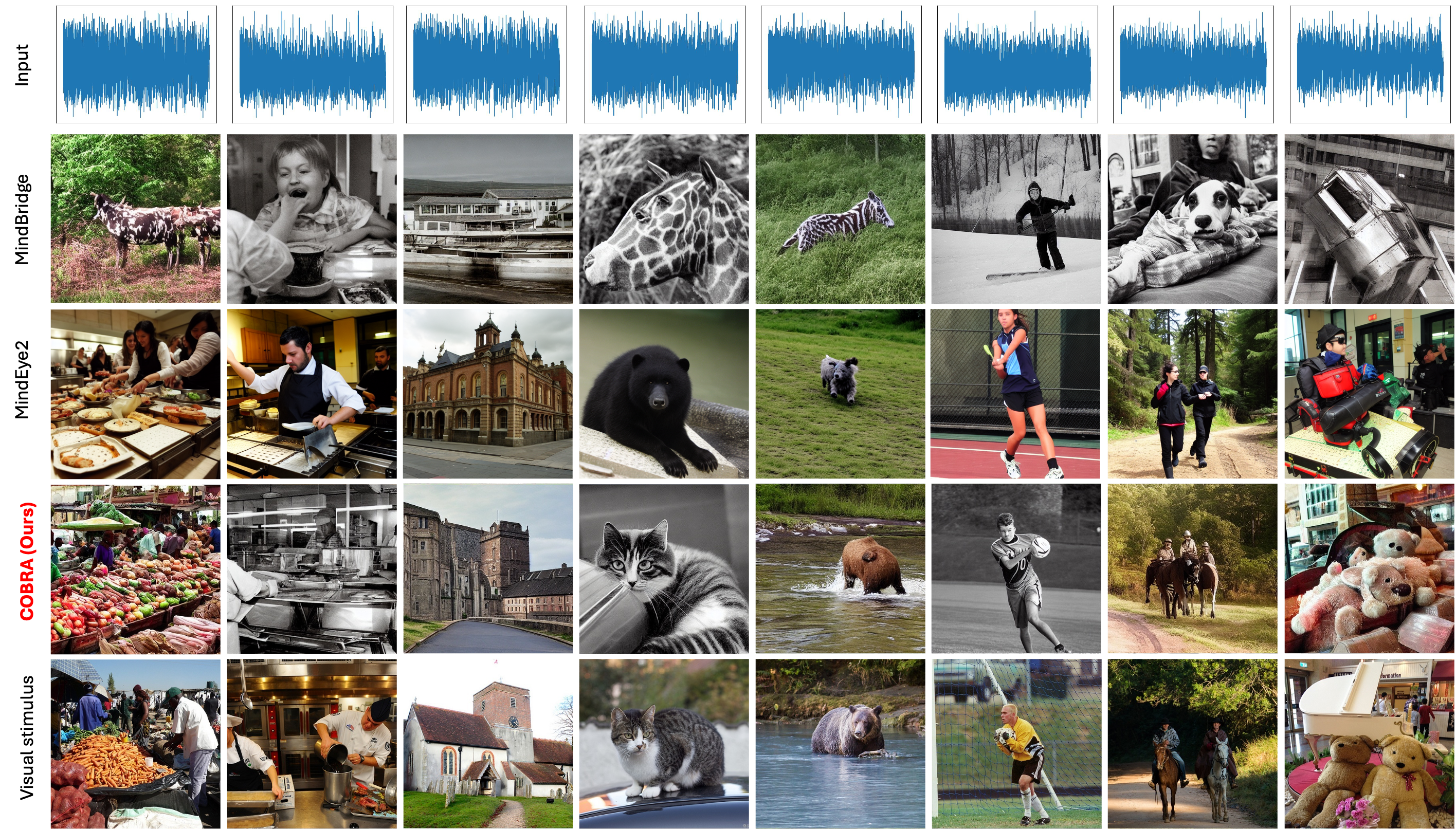}
    \caption{Qualitative comparison with the previous method on vision-brain reconstruction. 
    The first row is the input fMRI signals. The next three rows present the results of MindBridge \cite{mindbridge}, MindEye2 \cite{mindeye2}, and our COBRA.
    The last row is the stimulus.}
    \label{fig:qualitative_sota}
\end{figure*}

In addition to our CL setup, we report the results of vision-brain reconstruction using various ViT models for the SC module. As in Table \ref{tab:unified-model-comparison}, with the SC module using ViT-L, we achieve the best scores on all metrics except PixCorr and SSIM (ranking second in SSIM). These findings highlight two major points. First, COBRA not only achieves SOTA results in the CL setup but is also competitive in general vision-brain reconstruction. Second, COBRA’s performance is responsible for the depth of the SC module, indicating the framework’s scalability. Fig. \ref{fig:qualitative_sota} shows superior reconstruction results compared to prior work.

\subsection{Comparison To Previous SOTA Models of Brain Captioning}

\begin{figure*}
    \centering
    \includegraphics[width=0.85\linewidth]{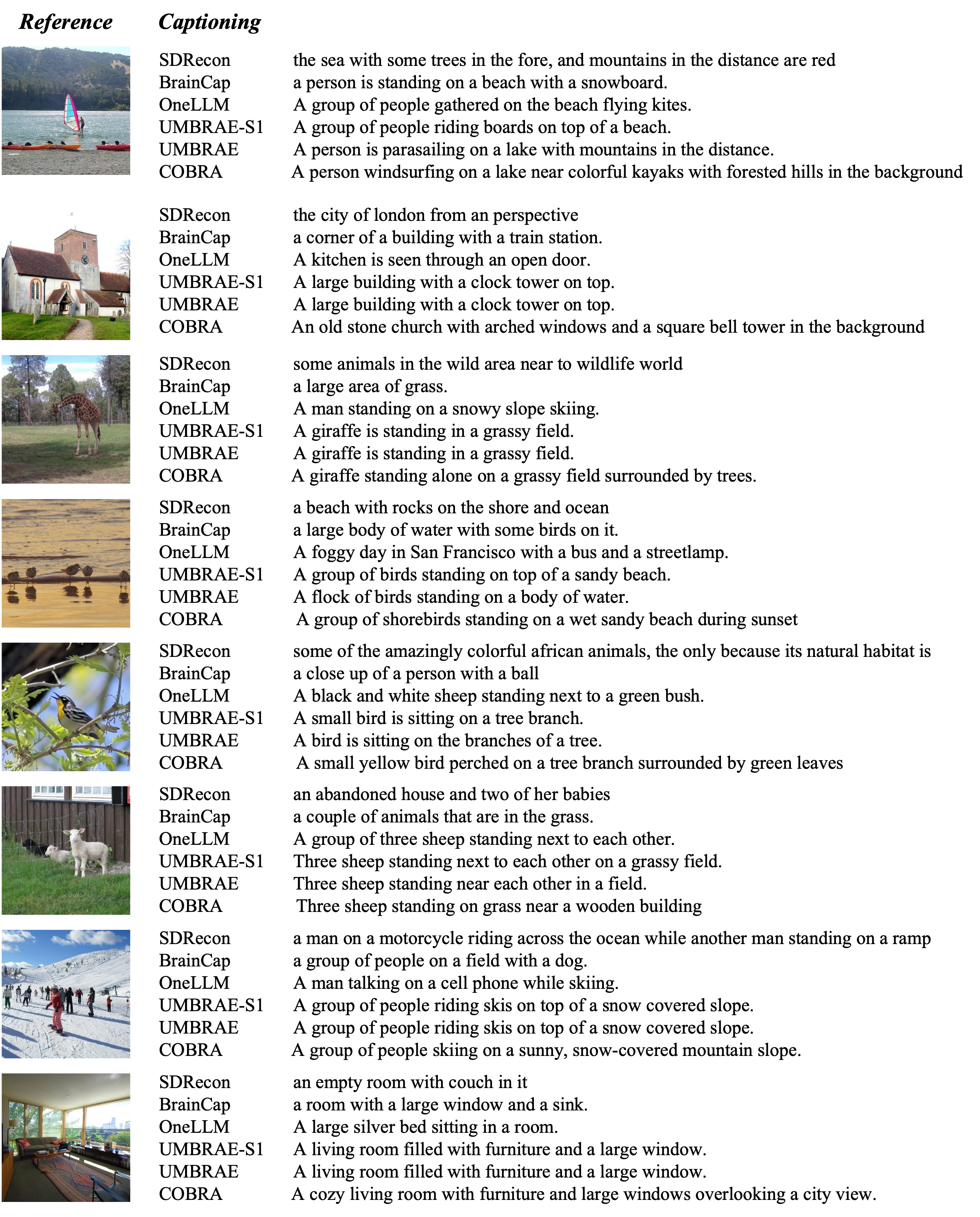}
    \caption{Qualitative comparison with the previous method on brain captioning task.}
    \label{fig:brain_captioning}
\end{figure*}

\BacRebuttal{
Following the experimental setup and evaluation protocol from \cite{xia-umbrae-2024}, we train our proposed COBRA model for the brain captioning task. As shown in Table~\ref{tab:brain_captioning}, COBRA consistently achieves state-of-the-art results across all standard captioning metrics, including BLEU-1 to BLEU-4, METEOR, ROUGE, CIDEr, SPICE, CLIP-S, and RefCLIP-S. Compared to prior approaches such as SDRecon~\cite{takagi2023high}, OneLLM~\cite{han2024onellm}, BrainCap~\cite{ferrante2023brain}, and UMBRAE~\cite{xia-umbrae-2024}, COBRA demonstrates substantial improvements, particularly in CIDEr and CLIP-S metrics, which are indicative of higher semantic alignment with the ground truth captions. Notably, COBRA outperforms UMBRAE, which is the current SOTA, by a clear margin in CIDEr (63.1 vs. 61.06) and CLIP-S (69.2 vs. 67.78), while also achieving the best scores on BLEU-4 and METEOR. This suggests that COBRA not only captures low-level lexical cues but also preserves the high-level semantics of the original stimuli more effectively .Fig. \ref{fig:brain_captioning} shows superior brain captioning results compared to prior work.
}

\begin{table}[t]
\caption{\textbf{Brain Captioning}. `Shikra-w/img' refers to the image captioning result from Shikra~\cite{chen2023shikra} using the ground truth image as input, serving as an approximate upper bound.
}
\setlength{\tabcolsep}{2pt}
\label{tab:brain_captioning}
\centering
\begin{tabular}{@{}lcccccccccc@{}}
\toprule
Method & BLEU1 & BLEU2 & BLEU3 & BLEU4 & METEOR & ROUGE & CIDEr  & SPICE  & CLIP-S & RefCLIP-S\\
\midrule
Shikra-w/img~\cite{chen2023shikra} & 82.38 & 69.90 & 58.63 & 49.66 & 35.60 & 65.49 & 161.43 & 27.62 & 80.60   & 85.92  \\
\midrule
SDRecon~\cite{takagi2023high}        & 36.21 & 17.11 & 7.72 & 3.43 & 10.03 & 25.13  & 13.83 & 5.02 & 61.07 & 66.36   \\
OneLLM~\cite{han2024onellm} & 47.04 & 26.97 & 15.49 & 9.51 & 13.55 & 35.05  & 22.99 & 6.26 & 54.80    & 61.28   \\
UniBrain~\cite{mai2023unibrain} &  -  & -  & - & - & {16.90}  & 22.20  &  - & -  & -  &  -  \\
BrainCap~\cite{ferrante2023brain} & {55.96} & {36.21} & {22.70} & {14.51} & 16.68 & {40.69} & {41.30} & {9.06} & {64.31} & {69.90} \\
UMBRAE-S1 \cite{xia-umbrae-2024}   & {57.63} & {38.02} & {25.00}  & {16.76}   & {18.41}   &  {42.15}  & {51.93} & {11.83} & {66.44} & {72.12}   \\
UMBRAE \cite{xia-umbrae-2024} & {59.44} & {40.48} & {27.66} & {19.03} & {19.45} & {43.71}  & {61.06} & \textbf{12.79} &  {67.78} & \textbf{73.54}  \\
\midrule
COBRA  & \textbf{61.2} & \textbf{42.4} & \textbf{30.1} & \textbf{19.4} & \textbf{21.7} & \textbf{44.3}  & \textbf{63.1} & 12.6 &  \textbf{69.2} & {73.5}  \\
\bottomrule
\end{tabular}
\end{table}

\subsection{Ablation Studies}

\textbf{Effectiveness of the vision-brain data representation}. We compared COBRA’s performance when using full brain signals versus partial ROI signals as input, as shown in Table \ref{tab:abl-data-feature-decoder}. Under identical settings, the model with full brain signals achieved 3\% higher in SSIM and 2\% higher in CLIP. It demonstrates that using full brain signals is significantly more effective than relying solely on partial ROI signals, as suggested by previous methods.

\noindent
\textbf{Effectiveness of the MRIFormer module}. In COBRA, our MRIFormer includes a transformer encoder to accumulate subject-specific and individual information, and a transformer decoder to translate fMRI data into CLIP space. 
As shown in Table \ref{tab:abl-data-feature-decoder}, our module outperforms the linear layers used in prior studies by approximately 3\% in both SSIM and CLIP, demonstrating the effectiveness of our approach. 

\noindent
\textbf{Effectiveness of prompt length for PSS module.} We evaluate the effectiveness of prompt length in the PSS module. In particular, we experiment with different top-$k$ values as in the Eqn \eqref{eq:specific_features}. The results are shown in Table \ref{tab:abl-prompt-length}. 
\BacRebuttal{We observed that $k=30$ consistently yields the best performance in both SSIM and CLIP metrics. Larger $k$ did not improve performance and increased model size; smaller $k$ resulted in underfitting the subject-specific space. Thus, $k=30$ balances expressiveness and efficiency.}

\begin{figure*}
    \centering
    \includegraphics[width=0.7\linewidth]{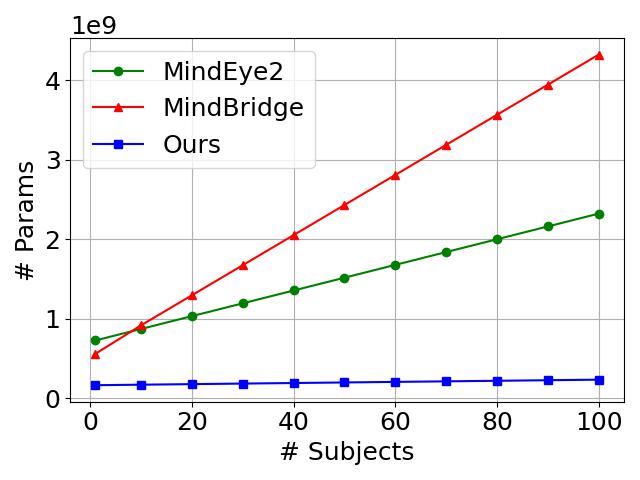}
    \caption{The size of model w.r.t number of subjects increasing.}
    \label{fig:model-size}
\end{figure*}

\noindent
\textbf{Model sizes during continual learning.} The size of COBRA grows gradually in response to an increasing number of subjects in CVBU. We simulate COBRA's size as the number of subjects increases from 1 to 100. 
Additionally, we compare its model size to that of MindBridge \ \cite{mindbridge} and MindEye2 \cite{mindeye2}, which are designed as unified models. The results are shown in Fig. \ref{fig:model-size}. Notably, the model size increases approximately nine times and three times for MindBridge and MindEye2, respectively, as the subject count rises from 1 to 100. Meanwhile, COBRA's size grows minimally. This is because only the PSS and MRIFormer modules were expanded to additional subjects. Designed as lightweight and simple modules, the PSS and MRIFormer module adds relatively few parameters compared to MindBridge and MindEye2.

\noindent
\textbf{Accuracy of SC and PSS Modules}
In the proposed COBRA, the SC module is designed to remain fixed when training on new subjects, as it captures common vision-brain patterns learned from previous subjects. To verify this, we trained the SC module on subjects {(3,4,6,8)} and tested it on the unseen subjects {(1,2,5,7)}. The SC module achieved an F1 score of 0.84, supporting our hypothesis. For the PSS module, we evaluated its accuracy in learning subject-specific features for each subject. \BacRebuttal{By using a prompt-based approach, } the PSS achieved a high accuracy of 98.5\%, indicating that the module effectively selects prompts that represent subject-specific features. \BacRebuttal{Furthermore, we also design the PSS module following attention pooling \cite{clip-paper-2021} or random-based approaches to select subject-specific tokens. The performance of these approaches is 82.8\% and 21.6\%, respectively. These results demonstrate the efficiency of our PSS module using a prompt-based approach.} The detailed results are presented in Table \ref{tab:abl-sc-pss-performance}.

\begin{figure}[!ht]
    \centering
    \includegraphics[width=1.0\linewidth]{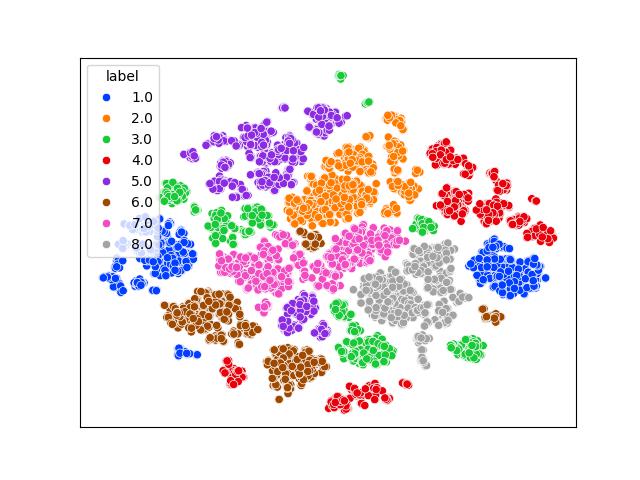}
    \caption{Visualization of the subject-specific features using T-SNE}
    \label{fig:tsne_ss_feature}
\end{figure}

\begin{table}[!ht]
\caption{Performance of SC and PSS module on their auxiliary tasks}
\label{tab:abl-sc-pss-performance}
\begin{tabular}{|l|l|l|r|}
\hline
Module & Approach & Metric & \multicolumn{1}{l|}{Accuracy} \\ \hline
SC & ViT & f1 score & 0.84 \\ \hline
PSS & Prompt-based & accuracy & 98.5 \\ \hline
PSS & Attention Pooling \cite{clip-paper-2021} & accuracy & 82.8 \\ \hline
PSS & Random & accuracy & 21.6 \\ \hline
\end{tabular}%
\end{table}
\noindent
\textbf{Visualization of Subject-Specific Features}
In this section, we analyze the discrimination of subject-specific features, i.e., $f_s$, that is extracted from the PSS module. We employ T-SNE \cite{van2008visualizing} to reduce the dimension of the features. The visualization is illustrated in Fig. \ref{fig:tsne_ss_feature}. The result demonstrates that the PSS module was able to learn the specific features that are discriminative for the subjects.

\begin{figure*}
    \centering
    \includegraphics[width=1.0\linewidth]{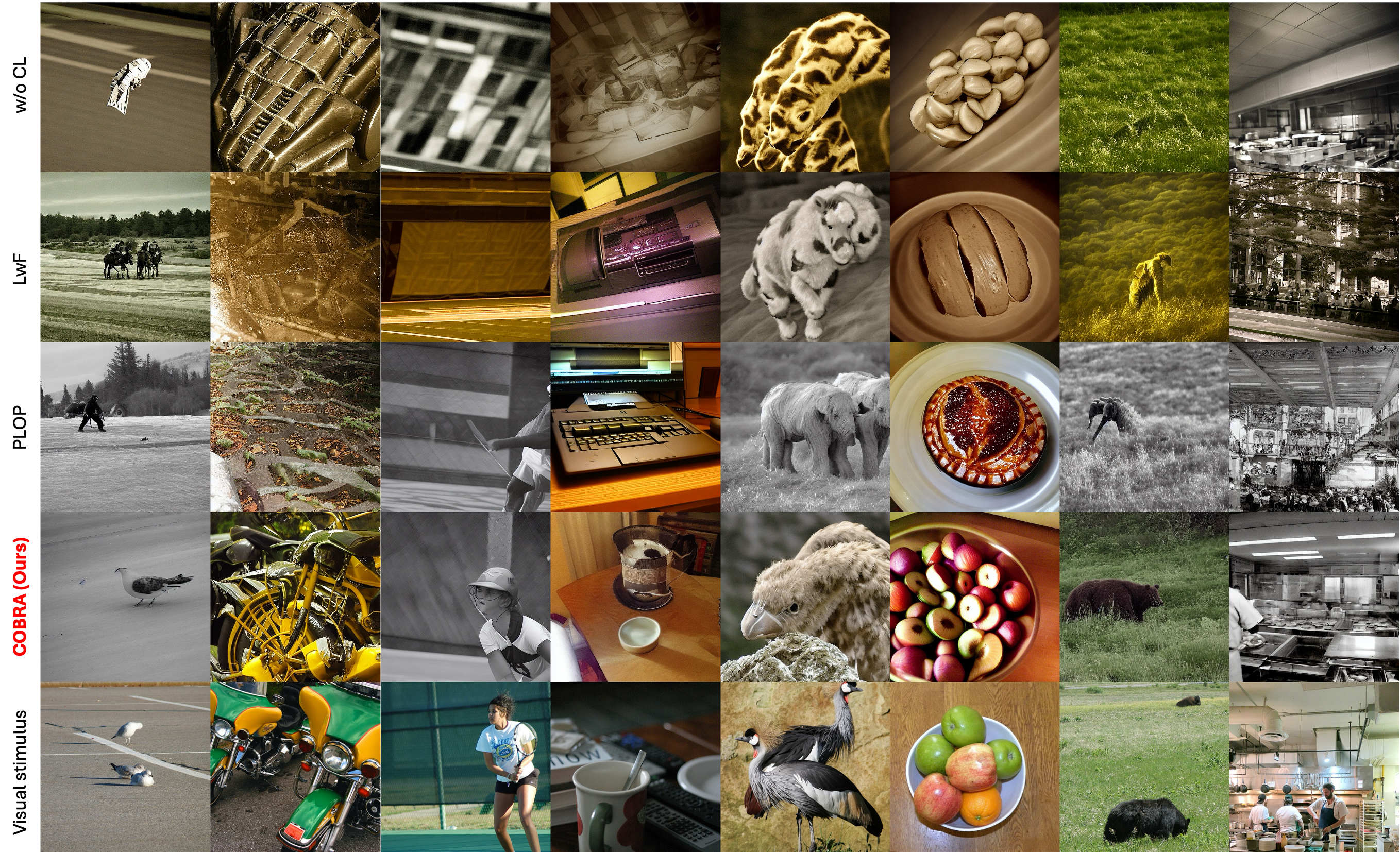}
    \caption{Qualitative comparison with the previous continual learning approaches on vision-brain reconstruction. From top to bottom, the first row shows the results of not using the continual learning approach. The next two rows are the results of LwF \cite{lwf} and PLOP \cite{plop}, respectively.
    The last two rows are the results of COBRA and visual stimulus, correspondingly.}
    \label{fig:cl_comparision_0}
\end{figure*}

\begin{figure*}
    \centering
    \includegraphics[width=1.0\linewidth]{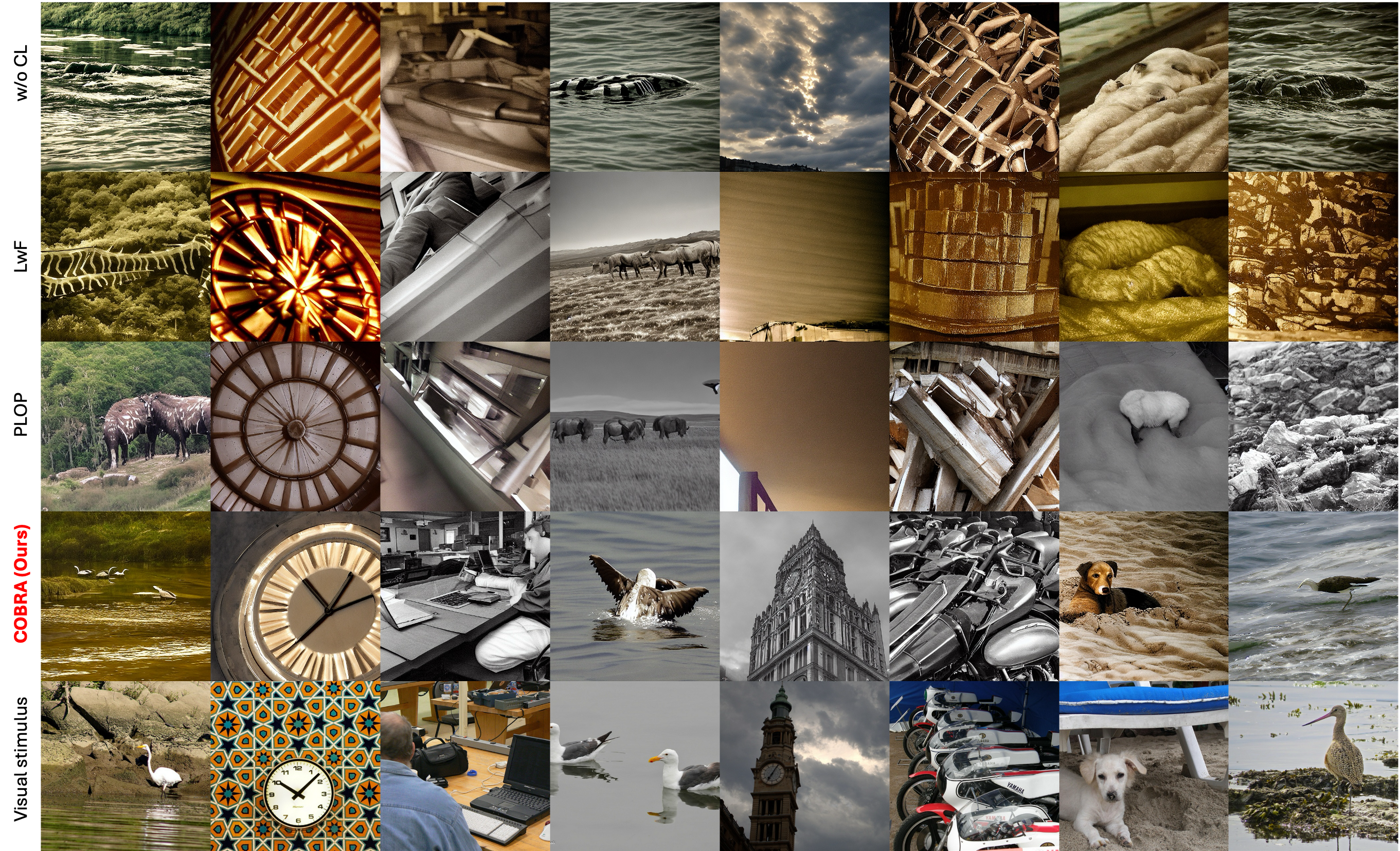}
    \caption{Qualitative comparison with the previous continual learning approaches on vision-brain reconstruction. From top to bottom, the first row shows the results of not using the continual learning approach. The next two rows are the results of LwF \cite{lwf} and PLOP \cite{plop}, respectively.
    The last two rows are the results of COBRA and visual stimulus, correspondingly.}
    \label{fig:cl_comparision_1}
\end{figure*}

\noindent
\textbf{Qualitative Comparison With Previous Continual Learning Approaches}. In this section, we compare the qualitative vision-brain reconstruction results of the model under four scenarios: without any continual learning (CL) approach, using LwF \cite{lwf}, using PLOP \cite{plop}, and our proposed method, COBRA. The results are illustrated in Fig. \ref{fig:cl_comparision_0} and Fig. \ref{fig:cl_comparision_1}. Our proposed method demonstrates significantly better and clearer reconstruction results compared to previous methods. In contrast, LwF \cite{lwf} and PLOP \cite{plop} lose critical contextual details. Notably, without applying any CL approach, the generated images are incoherent and fail to reconstruct the context meaningfully.

\noindent
\BacRebuttal{\textbf{Effect of Loss Balancing Hyperparameters}. We conducted an ablation study to evaluate the effect of loss weighting, with a focus on the regularization loss $\mathcal{L}{\text{reg}}$, which prevents prompt center collapse in the PSS module. As shown in Table~\ref{tab:lambda_ablation_full}, when $\lambda{\text{reg}}$ is removed ($\lambda_{\text{reg}}=0.0$), SSIM drops dramatically from 0.329 to 0.257 (a relative drop of over 21\%), and CLIP score falls from 92.9 to 88.4. This substantial degradation confirms that without sufficient regularization, the PSS prompt representations collapse to trivial solutions and fail to capture subject-specific variations.
In contrast, performance is less sensitive to variations in $\lambda_{\text{con}}$, which controls alignment with CLIP features. These results highlight the critical importance of $\mathcal{L}_{\text{reg}}$ in maintaining representation diversity and preventing catastrophic collapse, justifying our default configuration of $\lambda_{\text{reg}} = \lambda_{\text{con}} = 1.0$ in all experiments.
}

\begin{table}%
\caption{
    Ablation studies on using different input signals and the effectiveness of MRIFormer. 
    The bold numbers indicate the best results, while the underlines indicate the second-best results.
}
\label{tab:abl-data-feature-decoder}
\setlength{\tabcolsep}{1.5pt}
\begin{tabular}{|l|l|crrrrr|crrrrr|}
\hline
\multirow{3}{*}{Input Data} & \multirow{3}{*}{fMRI Trans.} & \multicolumn{6}{c|}{(3,4)-(6,8)-(1,2)-(5,7) (4 steps)} & \multicolumn{6}{c|}{(3,4,6,8)-(1,2,5,7) (2 steps)} \\ \cmidrule {3-14} 
 &  & \multicolumn{2}{c|}{(3,4,6,8)} & \multicolumn{2}{c|}{(1,2,5,7)} & \multicolumn{2}{c|}{all} & \multicolumn{2}{c|}{(3,4,6,8)} & \multicolumn{2}{c|}{(1,2,5,7)} & \multicolumn{2}{c|}{all} \\ \cmidrule {3-14} 
 &  & \multicolumn{1}{c|}{SSIM} & \multicolumn{1}{c|}{CLIP} & \multicolumn{1}{c|}{SSIM} & \multicolumn{1}{c|}{CLIP} & \multicolumn{1}{c|}{SSIM} & \multicolumn{1}{c|}{CLIP} & \multicolumn{1}{c|}{SSIM} & \multicolumn{1}{c|}{CLIP} & \multicolumn{1}{c|}{SSIM} & \multicolumn{1}{c|}{CLIP} & \multicolumn{1}{c|}{SSIM} & \multicolumn{1}{c|}{CLIP} \\ \hline
Partial ROI & Linear & \multicolumn{1}{r}{0.263} & 85.9 & 0.265 & 86.2 & 0.264 & 86.1 & \multicolumn{1}{r}{0.276} & 86.4 & 0.269 & 87.4 & 0.273 & 86.9 \\
Full Signals & Linear & \multicolumn{1}{r}{\underline{0.288}} & \underline{88.2} & \underline{0.291} & \underline{89.4} & \underline{0.290} & \underline{88.8} & \multicolumn{1}{r}{\underline{0.293}} & \underline{88.9} & \underline{0.297} & \underline{90.1} & \underline{0.295} & \underline{89.5} \\
Full Signals & MRIFormer & \multicolumn{1}{r}{\textbf{0.311}} & \textbf{93.2} & \textbf{0.344} & \textbf{93.1} & \textbf{0.328} & \textbf{93.2} & \multicolumn{1}{r}{\textbf{0.316}} & \textbf{92.3} & \textbf{0.342} & \textbf{93.4} & \textbf{0.329} & \textbf{92.9} \\
\hline
\end{tabular}%
\end{table}

\begin{table}[!b]
\caption{
    Ablation studies on different lengths $k$ of prompts. The bold numbers indicate the best results.
}
\label{tab:abl-prompt-length}
\setlength{\tabcolsep}{12pt}
{%
\begin{tabular}{|l|crrrrr|}
\hline
\multirow{3}{*}{top-$k$} & \multicolumn{6}{c|}{(3,4,6,8)-(1,2,5,7) (2 steps)} \\ %
\cmidrule {2-7}
& \multicolumn{2}{c|}{(3,4,6,8)} & \multicolumn{2}{c|}{(1,2,5,7)} & \multicolumn{2}{c|}{all} \\ 
\cmidrule {2-7} 
& \multicolumn{1}{c|}{SSIM} & \multicolumn{1}{c|}{CLIP} & \multicolumn{1}{c|}{SSIM} & \multicolumn{1}{c|}{CLIP} & \multicolumn{1}{c|}{SSIM} & \multicolumn{1}{c|}{CLIP} \\ \hline
10 & \multicolumn{1}{r}{0.301} & 90.9 & 0.331 & 92.0 & 0.316 & 91.5 \\
20 & \multicolumn{1}{r}{{0.304}} & {91.8} & {0.337} & {92.7} & {0.321} & {92.3} \\
30 & \multicolumn{1}{r}{\textbf{0.316}} & \textbf{92.3} & \textbf{0.342} & \textbf{93.4} & \textbf{0.329} & \textbf{92.9} \\
40 & \multicolumn{1}{r}{{0.312}} & {91.8} & {0.337} & {93.0} & {0.325} & {92.4} \\
\hline
\end{tabular}%
}
\end{table}

\begin{table}[!ht]
\centering
\caption{
Ablation study on the impact of loss balancing weights $\lambda_{\text{reg}}$ and $\lambda_{\text{con}}$ under the (3,4,6,8)-(1,2,5,7) continual learning setup. When $\lambda_{\text{reg}}$ is too small, the PSS module collapses, resulting in a substantial performance drop.
}
\label{tab:lambda_ablation_full}
\begin{tabular}{|c|c|c|c|}
\hline
$\lambda_{\text{reg}}$ & $\lambda_{\text{con}}$ & SSIM $\uparrow$ & CLIP $\uparrow$ \\
\hline
{0.0} & 1.0 & {0.257} & {88.4} \\
0.1 & 1.0 & 0.288 & 90.1 \\
0.5 & 1.0 & 0.318 & 92.5 \\
\textbf{1.0} & \textbf{1.0} & \textbf{0.329} & \textbf{92.9} \\
2.0 & 1.0 & 0.328 & 92.8 \\
\hline
1.0 & 0.5 & 0.324 & 92.1 \\
1.0 & 2.0 & 0.328 & 92.8 \\
\hline
\end{tabular}
\end{table}

\section{Conclusions}

This work has presented one of the first continual learning approaches to the Vision-Brain Understanding problems.
In particular, we have introduced the new Subject Commonality (SC) module to capture vision-brain patterns shared across subjects. Additionally, the Prompt-based Subject-Specific (PSS) module has been designed to learn unique vision-brain patterns for each subject. The MRIFormer is also incorporated to align fMRI features with consistent CLIP features. In this continual learning setup, only the PSS module needs updating to accommodate new subjects, effectively addressing catastrophic forgetting. Our COBRA approach has achieved state-of-the-art performance in the VBU tasks, as demonstrated on the NSD benchmark.

\begin{figure}[!t]
    \centering
    \includegraphics[width=1\linewidth]{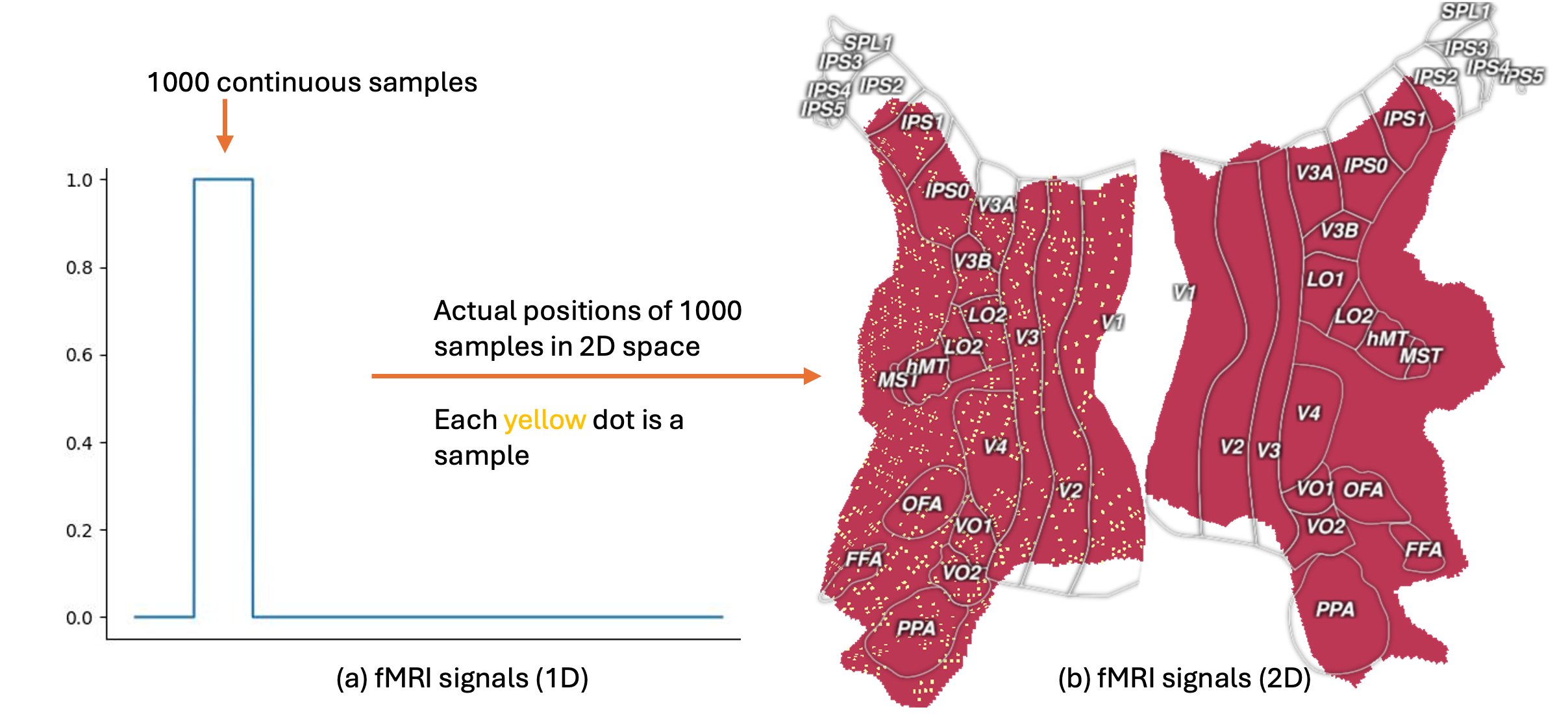}
    \caption{Demonstration of the spatial issue in 1D fMRI Signals.}
    \label{fig:spatial_issue}
\end{figure}

\noindent
\textbf{Limitations and Future Work}. 
Our SC module may be limited when the number of subjects is only one, as it does not effectively capture patterns shared across multiple subjects. 
The more subjects the module is trained on, the more accurately it can identify common patterns. 
In addition, this module is currently limited to 80 classes of MS-COCO. 
These limitations will motivate future research.

\noindent
\textbf{Broader Impact}. 
Advancements in deep learning and neuroscience hold transformative potential for enhancing adaptability in neuroscience applications. Implementing continual learning will empower researchers to identify common fMRI features based on a given condition in the future. It will facilitate more adaptable and sophisticated applications involving fMRI, such as the reconstruction of visual perception, decoding of internal speech, and development of neural prosthetics for communication.

\section{Data Availability}
The data that support the findings of this study are openly available at \href{https://naturalscenesdataset.org/}{https://naturalscenesdataset.org/}.

\section{Appendix}
\subsection{Explainations of Regions of Interest}
\Bac{
Regions of Interest (ROIs) are specific brain areas responsible for various perceptual functions in humans. For example, $floc-face$ and $floc-face$ are responsible for processing face and body while we are looking at someone. It means that the neurons in these regions are firing, indicating that the stimulus actively activates them. These regions are determined by experiments with statistics. For instance, consider the \textit{floc-faces} example. A set of visual stimuli is prepared to determine these regions, alternating between categories such as faces, objects, scenes, and words. fMRI signals are recorded from multiple participants as they are exposed to these stimuli. To pinpoint the \textit{floc-faces} regions, the brain's response to faces is contrasted with its responses to other categories (e.g., faces vs. objects) using statistical methods like General Linear Models (GLM).
}

\subsection{Explanation of Spatial Issue in 1D fMRI Signals}
\label{subsec:appendix_spatial_issue}
In this section, We provide additional details about the spatial issue discussed in Section \ref{subsec:revisit_data_representation} regarding fMRI signals represented in 1D form. Specifically, for a given subject's fMRI signals, we mask out all samples by setting their values to zero, retaining only \textit{1000 continuous samples} and assigning their values to one. These modified signals are then projected onto the fsaverage surface for visualization, as shown in Fig. \ref{fig:spatial_issue}. It is evident that in the 1D form, the \textit{1000 continuous samples} share strong correlations. However, when these signals are represented in 2D, the samples appear sparsely distributed. For this reason, analyzing fMRI signals in 1D form results in a significant loss of spatial information.

{
\bibliography{sn-bibliography}%
}

\end{document}